\documentclass[]{TEAI}

\usepackage{helvet}
\usepackage{amsmath}
\usepackage{amssymb}
\usepackage{amsfonts}
\usepackage[numbers,sort&compress]{natbib}
\usepackage{graphicx}
\usepackage{subcaption}
\usepackage[utf8]{inputenc}
\usepackage[T1]{fontenc}
\usepackage{hyperref}
\usepackage{url}
\usepackage{booktabs}
\usepackage{nicefrac}
\usepackage{microtype}
\usepackage{xcolor}
\usepackage{wrapfig}
\usepackage{titletoc}
\usepackage{tikz}
\usepackage{comment}
\usepackage{tabularx}
\usepackage{array}
\usepackage{etoolbox}
\usepackage{pgfplots}
\usepackage{float}
\usepackage{multirow}
\usepackage{makecell}
\usepackage{siunitx}
\usepackage{pgf-pie}
\usepackage[export]{adjustbox}
\usepackage{ragged2e}
\usepackage{caption}
\usepackage{enumitem}
\usepackage{pifont}
\usepackage[hang,flushmargin]{footmisc}
\usepackage{tcolorbox}
\usepackage{listings}
\usepackage{xspace}
\usepackage{longtable}

\tcbuselibrary{breakable}
\tcbuselibrary{skins}

\makeatletter
\DeclareRobustCommand\onedot{\futurelet\@let@token\@onedot}
\def\@onedot{\ifx\@let@token.\else.\null\fi\xspace}
\def\eg{\emph{e}\onedot g\onedot}

\def\ie{\emph{i}\onedot e\onedot}

\makeatother

\title{SegDiff: Segmented Trajectory Diffusion for Consistent and Adaptive Robot Manipulation}

\author[1,2]{Haidong Cao}
\author[1,2]{Wenjun Cao}
\author[1,2]{Quanhao Li}
\author[1,2]{Sicheng Xie}
\author[1,2]{Zhiying Du}
\author[1,2]{Jiaqi Leng}
\author[1,2,\dagger]{Zuxuan Wu}
\author[1,2]{Yu-Gang Jiang}

\affiliation[1]{Institute of Trustworthy Embodied AI, Fudan University, China}
\affiliation[2]{Shanghai Key Laboratory of Multimodal Embodied AI, China}
\contribution[\dagger]{Corresponding author}

\abstract{Imitation learning enables robots to acquire manipulation skills from demonstrations by mapping observations to actions. Existing approaches predict either short-horizon continuous action sequences or discrete keyposes. However, continuous prediction methods suffer from compounding errors due to short prediction horizons and struggle with multi-modal action distributions, whereas keypose-based methods necessitate an external planner, constraining real-time applicability. To address these challenges, we introduce \textbf{SegDiff}, a closed-loop visuomotor policy that integrates the strengths of both paradigms. SegDiff decomposes demonstrations into motion segments between keyposes and learns to predict the continuous trajectory from the current state to the next keypose, enabling long-horizon prediction with real-time refinement. Furthermore, we leverage the capability of diffusion models and DDIM inversion to propose a Dynamic Temporal Ensembling mechanism, which allows the policy to efficiently respond to dynamic environments and mitigate discontinuities caused by inconsistent multi-modal sampling. SegDiff demonstrates significant performance gains over existing approaches across various simulated and real-world scenarios, indicating its strong ability to reason over extended temporal dependencies while maintaining real-time adaptability and control stability.}

\begin{document}
\maketitle

\vspace{-1.5em}

\section{Introduction}
Robotic manipulation is a critical skill for intelligent agents, yet developing robust policies remains challenging due to the complexity of the physical world. Imitation learning (IL)~\cite{osa2018algorithmic, zare2024survey, schaal1996learning, schaal1999imitation, bain1995framework}, which learns directly from expert demonstrations, has thus emerged as a powerful and pragmatic paradigm. Offline imitation learning (OIL)~\cite{liu2021curriculum, bain1995framework, chen2020bail, arachchige2025sail} further improves data efficiency and safety by learning from fixed expert datasets without costly real-world interaction. Among OIL approaches, behavior cloning (BC)~\cite{bain1995framework} remains the most fundamental and widely used paradigm. Traditional BC predicts the immediate action from the current observation, but even small deviations between the predicted and expert actions caused by perception noise or imperfect policy approximation can quickly accumulate and drive the robot into out-of-distribution (OOD) states. 

\begin{figure}[t]
    \centering
    \includegraphics[width=\textwidth]{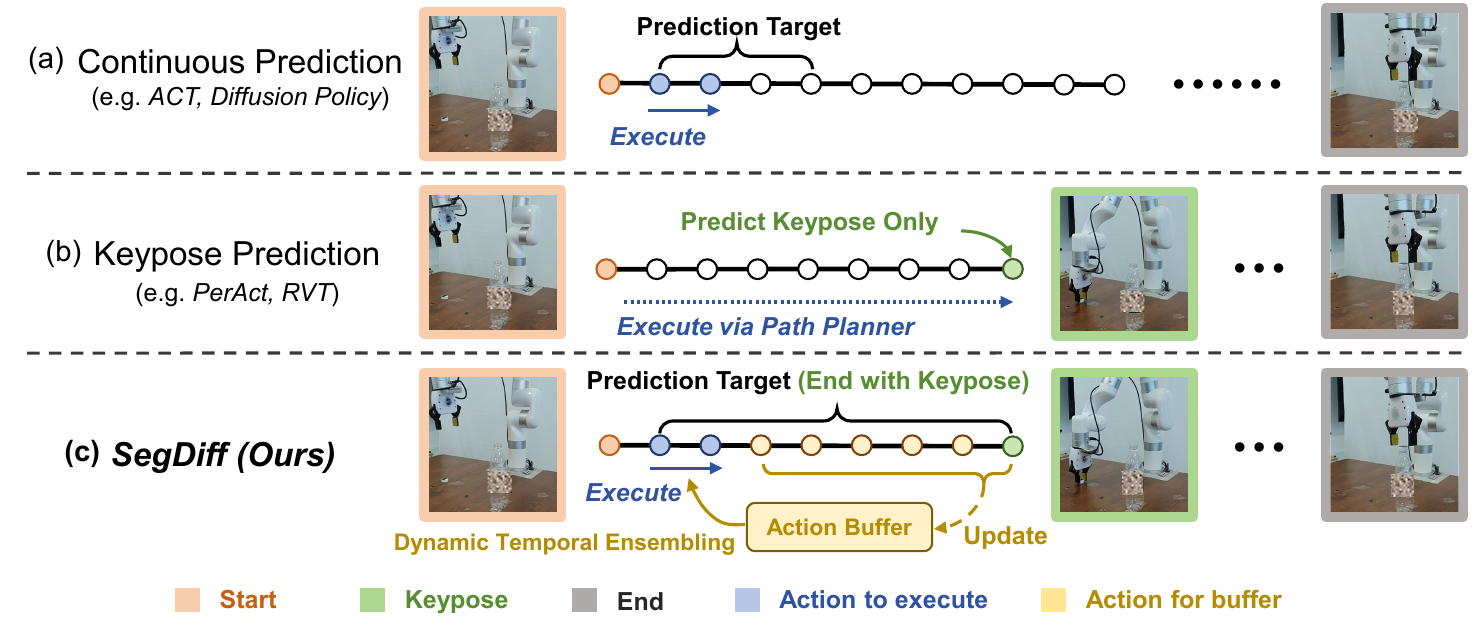}
    \caption{
        Comparison of conventional continuous prediction methods, keypose prediction methods, and \textbf{SegDiff}. SegDiff combines the advantages of both paradigms by enforcing keypose constraints within continuous trajectory generation via \textbf{Segmented Trajectory Modeling}, while improving action consistency and real-time responsiveness through \textbf{Dynamic Temporal Ensembling}.
    }
    \label{fig:overview}
\end{figure}

Recent approaches~\cite{chi2025diffusion, zhao2023learning, ze20243d, black2024pi_0, black2025real, team2024octo, kim2025fine, ze2025generalizable} extend the prediction horizon by predicting short action sequences as shown in Fig.~\ref{fig:overview}(a), with some employing temporal ensembling~\cite{zhao2023learning, black2025real} between consecutive predictions to smooth motions. Although these methods improve short-term consistency, they still suffer from compounding errors over extended durations. Another major challenge arises from the multi-modal nature of the action distribution in robotic manipulation tasks. The same observation can correspond to multiple valid action modes such as grasping an object from different directions. Averaging across such modes can yield ambiguous actions and cause unsafe behaviors. 

Keypose-based prediction methods~\cite{james2022q, shridhar2023perceiver, ze2023gnfactor, goyal2023rvt, ke20253d, goyal2024rvt, fang2025sam2act} mitigate compounding errors by focusing on discrete salient points (\ie, keyposes) and applying an external planner to generate motions toward the next keypose in an open-loop manner, as shown in Fig.~\ref{fig:overview}(b). However, this open-loop design limits the adaptability to dynamic environments and constrains real-time applicability. While there are approaches that attempt to combine continuous action prediction and keypose prediction~\cite{yu2024bikc, wang2025hierarchical, zhang2025chain}, they often rely on complex multi-stage architectures that increase inference latency and suffer from error propagation between modules. Furthermore, they typically struggle to handle the multi-modal nature of robotic actions, making them susceptible to mode switching and limiting their responsiveness in dynamic scenarios.

To address these challenges, we propose \textbf{SegDiff}, a closed-loop robot manipulation policy that seamlessly integrates the strengths of continuous prediction and keypose prediction. The core idea is to leverage keyposes as \textbf{prediction anchors} and learn continuous motion to reach the next keypose. Specifically, we formulate the problem as predicting the complete trajectory from the current state to the next keypose, which we term \textbf{Segmented Trajectory Modeling} (as shown in Fig.~\ref{fig:overview}(c)). This formulation directs learning toward task-critical states and alleviates the compounding errors common in continuous prediction.

During inference, SegDiff follows the Receding Horizon Control (RHC)~\cite{mayne1988receding} scheme, executing only the initial portion of each predicted trajectory and storing the remainder in an action buffer. Benefiting from Segmented Trajectory Modeling, the buffer and the newly predicted trajectory share the same initial state and both terminate at the next keypose. This inherent alignment improves the accuracy of next keypose prediction and enables detection of action mode deviations. Leveraging this, SegDiff employs a \textbf{Dynamic Temporal Ensembling} mechanism using DDIM inversion to validate and refine the action buffer.
Outdated buffers are discarded and replaced by trajectories generated conditioned on the latest observation. Conversely, when the buffer remains valid, new trajectories are generated from the latest observation while preserving the original action mode, which facilitates more effective temporal ensembling with the existing buffer. This entire design enables SegDiff to adapt to dynamic environments while ensuring modal consistency across consecutive predictions. In summary, our contributions are as follows:

1) We propose \textbf{SegDiff} (Segmented Trajectory Diffusion), featuring a novel \textbf{Segmented Trajectory Modeling} approach that integrates keyposes as prediction anchors. This formulation not only serves as a training objective to guide the model toward task-critical states, but also acts as a crucial architectural enabler, laying the structural foundation for real-time action buffer refinement via DDIM inversion.

2) Building upon this structural design, we introduce a novel \textbf{Dynamic Temporal Ensembling} mechanism. By leveraging DDIM inversion, this mechanism enables the policy to seamlessly adapt to dynamic environments while strictly maintaining action mode consistency across consecutive predictions.

3) We demonstrate the superior and robust performance of SegDiff on two simulated benchmarks and five challenging real-world manipulation tasks, showing that SegDiff consistently outperforms existing advanced approaches.

\section{Related Work}
\label{sec:relatedintro}
\noindent\textbf{Keypose-Based Manipulation.} Keypose-based policies reformulate the robotic manipulation task into predicting the next end-effector pose. Early methods~\cite{james2022q, james2022coarse, shridhar2023perceiver} relied on pixel-level or voxel-level representations to locate the keypose, but incurred high computational cost and lacked real-time performance. Subsequent methods~\cite{chen2023polarnet, ze2023gnfactor, gervet2023act3d, goyal2024rvt, fang2025sam2act, ke20253d} adopt representations such as neural fields or point clouds for improved spatial precision, yet still depend on external planners~\cite{karaman2011sampling, karaman2011anytime, lavalle1998rapidly, kuffner2000rrt} and struggle in dynamic environments. ChainedDiffuser~\cite{xian2023chaineddiffuser} utilizes a diffusion model to generate intermediate trajectories, but its two-stage inference pipeline limits real-time adaptability. In contrast, our method predicts the complete trajectory from the current pose to the next keypose at every inference step, enabling real-time adaptation while preserving the keypose constraint.

\vspace{0.05in}
\noindent\textbf{Continuous Action Prediction.} Continuous prediction methods~\cite{chi2025diffusion, zhao2023learning, ze20243d, black2024pi_0, team2024octo, kim2025fine, liu2025rdt, prasad2024consistency, lu2024manicm,xie2026human2robot, hu2025video} operate under a sliding-window paradigm, predicting short-horizon action sequences at each step while paying limited attention to task-critical states, which makes them prone to compounding errors. Waypoint-based and stage-conditioned methods~\cite{shi2023waypoint, sundaresan2025s, belkhale2023hydra, wang2025hierarchical, zhang2025chain, yu2024bikc} partially address this issue, but require extra annotations or incur multi-stage inference overhead. In contrast, our approach segments demonstrations using keyposes and predicts continuous trajectories toward the next keypose within a single end-to-end model.

\vspace{0.05in}
\noindent\textbf{Action Consistency and Temporal Ensembling.}
Temporal ensembling improves the smoothness of continuous predictions by aggregating outputs over time; however, abrupt mode switches can still produce unsafe actions. Recent approaches address this issue by adjusting denoising schedules~\cite{hoeg2024streaming, chen2025responsive, duan2025real} or incorporating optimization for consistency~\cite{black2025real}. However, these methods typically assume that previous predictions remain reliable, which may not hold in dynamic environments. SegDiff addresses this challenge with Dynamic Temporal Ensembling, which leverages DDIM inversion to explicitly validate and refine the action buffer in real time, improving both mode consistency and responsiveness.
\section{Method}
\label{sec:method}

\subsection{Preliminaries}
\label{sec:method-bg}
\noindent\textbf{Offline Imitation Learning.} We focus on offline imitation learning (OIL), where the objective is to learn a visuomotor policy solely from expert demonstrations without interaction with the environment. In this setting, the demonstration dataset is denoted as $\mathcal{D} = \left\{ \left( o_t, s_t, a_t \right) \right\}_{t=1}^{N}$, where $o_t$ represents the visual observation (\eg, RGB image) at time step $t$, $s_t$ denotes the robot state (\eg, end-effector pose), and $a_t$ is the expert action. The policy $\pi_\theta\left( a_t \mid o_t, s_t \right)$ is typically trained with the objective:
\begin{equation}
    \mathcal{L}_{\text{BC}} = \mathbb{E}_{(o_t, s_t, a_t) \sim \mathcal{D}} \left[ \| a_t - \pi_\theta \left( o_t, s_t \right) \|^2 \right].
\end{equation}
OIL methods can be broadly categorized into continuous prediction and keypose prediction. Continuous prediction methods replace the target $a_t$ with a future action sequence $a_{t:t+L}$, improving long-horizon reasoning. During inference, they typically follow the Receding Horizon Control (RHC) scheme~\cite{mayne1988receding}, executing part of the predicted sequence and re-planning with new observations. This setup naturally supports temporal ensembling~\cite{zhao2023learning}, where the unexecuted portion of the previous sequence is blended with the latest prediction for more accurate and smoother actions.
A representative approach is Diffusion Policy (DP)~\cite{chi2025diffusion}, which models the conditional distribution of continuous action sequences using a diffusion process~\cite{ho2020denoising}. Its training objective is formulated as:
\begin{equation}
    \mathcal{L}_\text{DP} = 
    \mathbb{E}_{a^0_{t:t+l},\, \epsilon,\, k}
    \left[
    \| \epsilon - \epsilon_\theta\left( a^k_{t:t+l}, k, o_{t-h:t}, s_t \right) \|^2
    \right],
\end{equation}
where $k$ denotes the diffusion step, $l$ is the length of the predicted action sequence, and $h$ indicates the number of historical observations used. 

Keypose prediction methods~\cite{james2022q, shridhar2023perceiver, ze2023gnfactor, gervet2023act3d, goyal2024rvt, ke20253d}, by contrast, replace $a_t$ with the next keypose (\eg, grasp pose), focusing on the most salient points within a trajectory while omitting intermediate motions. 

\noindent\textbf{DDIM Inversion. } To enable prior-informed trajectory refinement within the diffusion policy framework, we employ DDIM~\cite{song2021denoising}, which provides a deterministic formulation of diffusion models that permits inversion of the sampling process. The deterministic form of DDIM~\cite{song2021denoising} modifies the sampling process in DDPM~\cite{ho2020denoising} to:
\begin{equation}
\scalebox{0.9}{$
    x_{t-1}\! =\! \sqrt{\bar\alpha_{t-1}} \left( \frac{x_t - \sqrt{1-\bar\alpha_t}\,\epsilon_\theta\left( x_t, t \right)}{\sqrt{\bar\alpha_t}} \right) + \sqrt{1-\bar\alpha_{t-1}}\,\epsilon_\theta\left( x_t, t \right)
    $}
\end{equation}
This deterministic process enables \textbf{DDIM inversion}, which maps a data point $\mathbf{x}_0$ to its corresponding noise vector $\mathbf{x}_K$ for prior-informed sampling and conditional generation.

\begin{figure*}[t]
  \centering
  \includegraphics[width=\linewidth]{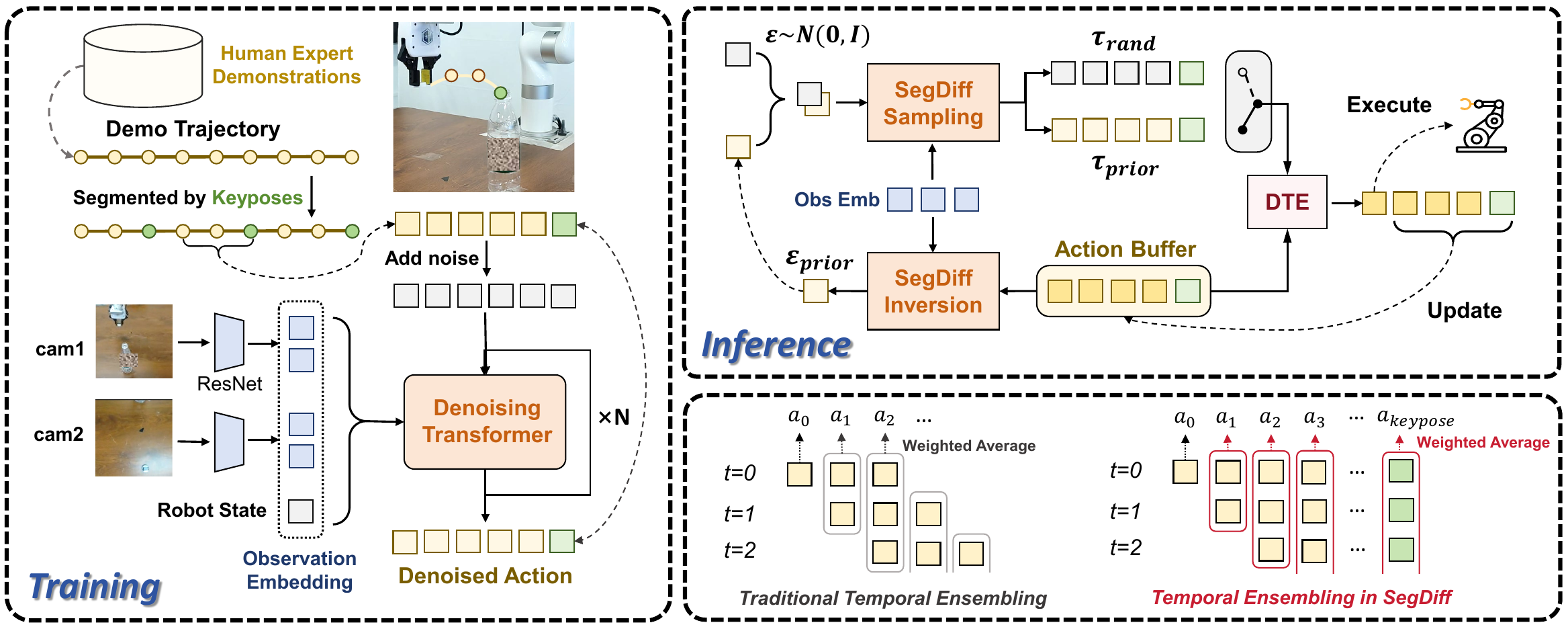}
  \caption{\textbf{Overview of SegDiff}. During training, demonstration episodes are segmented by keyposes, and the model learns to predict the complete trajectory from the current state to the next keypose (\textbf{Segmented Trajectory Modeling}). During inference, \textbf{DDIM inversion}~\cite{song2021denoising} generates a prior-informed candidate, allowing \textbf{Dynamic Temporal Ensembling (DTE)} to produce a trajectory that adapts to the current observation while maintaining temporal consistency for execution and buffer update.}
  \label{fig:framework}
\end{figure*}

\subsection{Segmented Trajectory Modeling}
\label{sec:method-stm}

As discussed in Sec.~\ref{sec:method-bg}, continuous prediction models generate smooth actions but suffer from compounding errors inherent to their sliding-window paradigm. In contrast, keypose-based methods focus on the most important states in a demonstration (keyposes), making them particularly suitable for precise manipulation tasks; however, their open-loop nature requires external planners and prevents real-time adaptation. While approaches such as ChainedDiffuser~\cite{xian2023chaineddiffuser} attempt to model intermediate paths, they remain limited in dynamic scenarios. 

To bridge these paradigms, we introduce \textbf{Segmented Trajectory Modeling} (STM). By using keyposes as prediction anchors, SegDiff mitigates compounding errors (see the supplementary material for a formal derivation) while maintaining a closed-loop control paradigm. Specifically, instead of predicting isolated keyposes or fixed-horizon action sequences, the model is trained to predict a continuous trajectory that originates from an arbitrary timestep $t$ and terminates at the subsequent keypose. 
We follow PerAct~\cite{shridhar2023perceiver} to extract keyposes from expert demonstrations in which a keypose is defined as where the gripper state changes (\ie, opening or closing) or the velocity of the end-effector approaches zero. We denote a specific human expert demonstration trajectory as $\left\{ a_1, a_2, \dots, a_T \right\}$, and extracted keyposes as $\left\{ a_{k_1}, a_{k_2}, \dots, a_{k_n} \right\}$. To ensure complete coverage of the entire trajectory, the last action $a_T$ is included in the keypose sequence $\left( a_{k_n} = a_T \right)$, and we define $a_{k_0} = a_1$. The trajectory can then be segmented into multiple sub-trajectories based on these keyposes. 
Each sample in the training dataset is therefore represented as:

\begin{equation}
\left\{ \left( o_{t-h:t},\, s_t,\, a_{t:k_m} \right)
\mid k_{m-1} \le t < k_m \right\},  
\end{equation}
where $ 1 \le m \le n$, and $h$ denotes the duration of the observation history. In this way, we obtain demonstration sub-segments starting from an arbitrary timestep $t$ and ending at the subsequent keypose $a_{k_m}$. Utilizing these sub-segments as training samples ensures that our model focuses on the keyposes of the demonstrations while simultaneously retaining an understanding of the intermediate motion process to support continuous action prediction.

We use a diffusion model to learn the conditional distribution of these sub-segments. Because each sub-segment has a variable length $k_m - t$, we interpolate all trajectories to a fixed length $L$, using cubic spline interpolation for translation and spherical linear interpolation for rotation. The training objective for the diffusion policy then becomes:
\begin{equation}
    \mathcal{L}_\text{SegDiff} = 
    \mathbb{E}_{a_{t:k_m},\, \epsilon,\, k}
    \left[
    \| \epsilon - \epsilon_\theta\left( \tilde{a}^{k}_{\left( t,k_m \right)}, k, o_{t-h:t}, s_t \right) \|^2
    \right],
\end{equation}
where $\tilde{a}_{\left( t,k_m \right)}$ is the sub-segment of fixed length $L$ obtained by interpolating the ground-truth sub-segment $a_{t:k_m}$, and $k$ denotes the diffusion timestep sampled uniformly from $\left\{ 1,2,\dots,K \right\}$.

\subsection{Dynamic Temporal Ensembling for Inference}
\label{sec:method-dte}

\begin{figure*}[t]
    \centering
    \includegraphics[width=\linewidth]{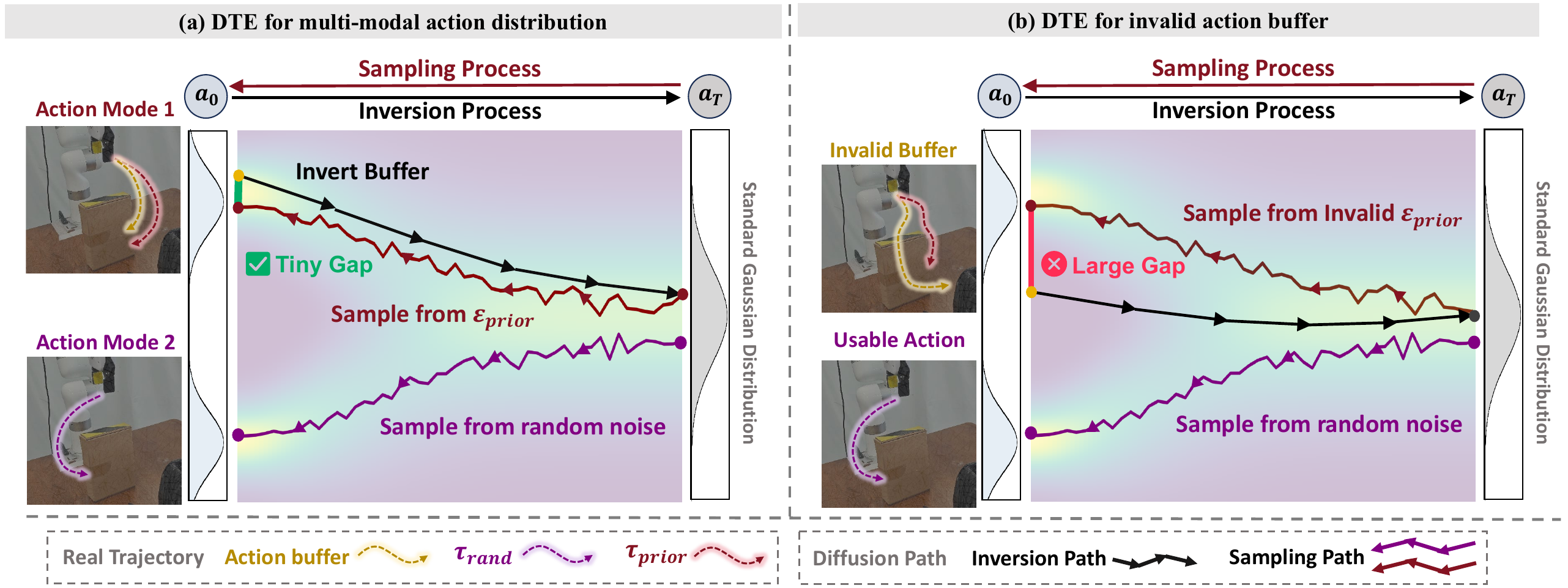}
    \caption{Comparison between sampling from noise inverted from action buffer and from random noise. If there exists a large gap between action buffer and the action sampled from the inverted noise, we discard the action buffer as unusable.}
    \label{fig:ddim}
\end{figure*}

\begin{figure}[t]
    \centering
    \includegraphics[width=0.8\linewidth]{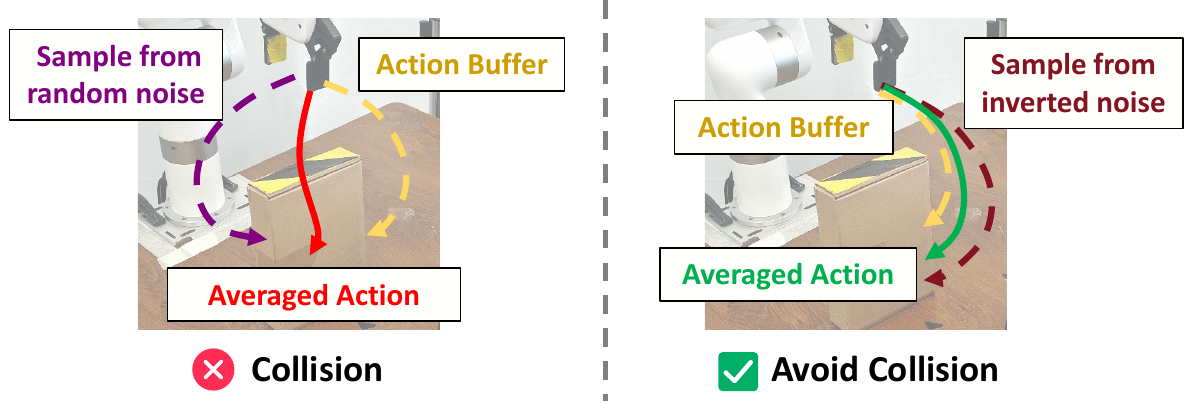}
    \caption{Mode switching in model predictions can lead to undesirable actions (left), whereas sampling from the inverted noise provides a consistent alternative (right).}
    \label{fig:colli}
\end{figure}

During inference, SegDiff follows the RHC scheme (Sec.~\ref{sec:method-bg}). At step $t$, the policy predicts a trajectory $\mathbf{\tau}_{t} = \{a_t, \dots, a_{t+L-1}\}$, executing only the first $n_{exec}$ actions while storing the remainder in an \textbf{action buffer} $b_t$ for temporal ensembling with later predictions. However, naive reuse of the action buffer introduces two challenges. First, due to the multi-modality in action distributions (Fig.~\ref{fig:ddim}(a)), newly predicted trajectories may belong to different modes, and directly switching between modes can lead to unsafe behavior (Fig.~\ref{fig:colli}, left). Second, if environmental changes invalidate the action buffer, averaging the invalid buffer with the latest prediction may cause the robot to respond too slowly to environmental changes and even enter out-of-distribution states, potentially resulting in task failure.

To address these challenges, we propose \textbf{Dynamic Temporal Ensembling} (DTE), a mechanism that leverages DDIM inversion to validate and refine the action buffer in real time. Unlike standard temporal ensembling which operates on a \textit{sliding window} with shifting horizons, SegDiff instead anchors trajectory prediction to the next keypose (Sec.~\ref{sec:method-stm}). Consequently, the action buffer $b_t$ and the newly predicted trajectory $\mathbf{\tau}_{t+1}$ share the same starting point and both terminate at the next keypose. This alignment facilitates more accurate keypose refinement as the keypose is repeatedly predicted before execution and provides the basis for using DDIM inversion to assess buffer consistency.

At each inference step $t$, we generate two candidates: (1) a \textbf{standard prediction} $\mathbf{\tau}_{\text{rand}}$ sampled from Gaussian noise, representing the most responsive prediction; and (2) a \textbf{prior-informed prediction} $\mathbf{\tau}_{\text{prior}}$, obtained by first performing DDIM inversion on the action buffer $b_t$ to acquire a noise vector $\mathbf{\epsilon}_{\text{prior}}$, and then denoising from $\mathbf{\epsilon}_{\text{prior}}$ conditioned on the new observation. As illustrated in Fig.~\ref{fig:ddim}(a), if the action buffer is valid, denoising from the DDIM-inverted noise $\mathbf{\epsilon}_{\text{prior}}$ (red sampling path) under the current observation yields a trajectory that remains consistent with the action mode of the buffer. Conversely, if the action buffer is no longer feasible under the updated observation, sampling from the inverted noise produces a trajectory that deviates significantly from $b_t$ (Fig.~\ref{fig:ddim}(b)).

We quantify this mode-level consistency and feasibility using two RMSE-based metrics. The \textbf{buffer validity} is defined as 
$d_{\text{valid}} = \mathrm{RMSE}\left( b_t, \mathbf{\tau}_{\text{prior}} \right)$, 
where a large $d_{\text{valid}}$ indicates that the buffered actions are no longer feasible under the current observation. 
The \textbf{mode discrepancy} is defined as 
$d_{\text{mode}} = \mathrm{RMSE}\left( \mathbf{\tau}_{\text{rand}}, \mathbf{\tau}_{\text{prior}} \right)$, 
where a large $d_{\text{mode}}$ indicates that the standard prediction deviates from the buffered action mode. 
To determine whether these deviations are significant, we introduce a threshold $\delta$. 
The threshold $\delta$ is determined in a data-driven manner by computing the DDIM inversion reconstruction error on the training set and selecting the maximum observed RMSE.

Formally, the Dynamic Temporal Ensembling (DTE) mechanism computes the final action sequence $\mathbf{\tau}_{DTE}$ as:

\begin{equation}
\mathbf{\tau}_{DTE} = 
\begin{cases}
\eta \cdot \mathbf{\tau}_{\text{rand}} + \left( 1-\eta \right) \cdot b_{t}, & d_{valid} \leq \delta \text{ and } d_{mode} \leq \delta \\
\eta \cdot \mathbf{\tau}_{\text{prior}} + \left( 1-\eta \right) \cdot b_{t}, & d_{valid} \leq \delta \text{ and } d_{mode} > \delta \\
\mathbf{\tau}_{\text{rand}}. & d_{valid} > \delta
\end{cases}
\label{eq:dte}
\end{equation}
Here, $\eta \in [0,1]$ is an update factor that controls the blending ratio between the new prediction and the action buffer. 

The rule in Eq.~\ref{eq:dte} operates as follows. 
When the buffer is valid ($d_{\text{valid}} \leq \delta$) and the new prediction remains mode-consistent ($d_{\text{mode}} \leq \delta$), we prioritize $\mathbf{\tau}_{\text{rand}}$ due to its stronger conditioning on the latest observation while blending it with the buffer. If the buffer is valid but a mode switch is detected ($d_{\text{mode}} > \delta$), we suppress the switch by utilizing $\mathbf{\tau}_{\text{prior}}$ to preserve action-mode consistency. In contrast, when the buffer becomes invalid ($d_{\text{valid}} > \delta$), it is discarded entirely and $\mathbf{\tau}_{\text{rand}}$ is fully adopted to rapidly adapt to environmental changes.

The resulting $\mathbf{\tau}_{DTE}$ is then used for execution and buffer update: the first $n_{exec}$ actions are executed, and the remaining trajectory is interpolated to length $L$ to form $b_{t+1}$. 
This cycle repeats until the end-effector reaches the target keypose within a predefined tolerance $\delta_r$. Regarding computational efficiency, the DDIM inversion step count is decoupled from the sampling process, which enables flexible configurations that balance efficiency and precision. For example, lightweight inversion can be combined with multi-step sampling to maintain trajectory accuracy without significant computational overhead (analyzed in Sec.~\ref{sec:ablation-time}). In addition, the sampling of $\mathbf{\tau}_{\text{rand}}$ and $\mathbf{\tau}_{\text{prior}}$ can be performed in parallel, resulting in negligible additional latency.

\begin{table}[!b]
  \begin{center}
  \setlength{\tabcolsep}{6pt}
  \caption{Success rates (\%) on 10 RLBench tasks~\cite{james2020rlbench}. Results for ACT, DP, CoA and Octo are taken from ~\cite{zhang2025chain} and ~\cite{zhang2025effective}. We also reproduce the results of DP ourselves and report them as DP*. The results of SegDiff are averaged over 3 random seeds.}
  \label{tab:rlbench}
  \begin{tabular}{@{}lcccccc@{}}
    \toprule
    Task 
    & ACT~\cite{zhao2023learning}
    & DP~\cite{chi2025diffusion}
    & DP*~\cite{chi2025diffusion}
    & CoA~\cite{zhang2025chain}
    & Octo~\cite{zhang2025effective}
    & SegDiff \\
    \midrule
    
    Sweep Dust & \textbf{100.0} & \textbf{100.0} & \textbf{100.0} & 92.0 & 80.0 & \textbf{100.0}\scriptsize{$\pm$0.0} \\
    Push Button & 8.0 & 12.0 & 24.0 & 76.0 & 76.0 & \textbf{90.7}\scriptsize{$\pm$2.3} \\
    Stack Wine & 56.0 & 56.0 & 56.0 & 80.0 & 52.0 & \textbf{94.7}\scriptsize{$\pm$2.3} \\
    Turn Tap & 36.0 & 32.0 & 32.0 & 56.0 & 28.0 & \textbf{74.7}\scriptsize{$\pm$8.3} \\
    Open Drawer & 52.0 & 44.0 & 52.0 & 88.0 & 84.0 & \textbf{97.3}\scriptsize{$\pm$2.3} \\
    Pick Up Cup & 20.0 & 0.0 & 76.0 & 80.0 & 44.0 & \textbf{90.7}\scriptsize{$\pm$2.3} \\
    Take Lid & 40.0 & 60.0 & 84.0 & 80.0 & 76.0 & \textbf{90.7}\scriptsize{$\pm$2.3} \\
    Press Switch & 52.0 & 56.0 & 28.0 & 44.0 & 44.0 & \textbf{68.0}\scriptsize{$\pm$4.0} \\
    Reach Target & 88.0 & 8.0 & 72.0 & 84.0 & 60.0 & \textbf{97.3}\scriptsize{$\pm$2.3} \\
    Open Box & 36.0 & 48.0 & 20.0 & 76.0 & \textbf{96.0} & 62.7\scriptsize{$\pm$6.1} \\
    \midrule
    \textbf{Average} & 48.8 & 41.6 & 54.4 & 75.6 & 64.4 & \textbf{86.7}\scriptsize{$\pm$1.4} \\
    \bottomrule
  \end{tabular}
  \end{center}
\end{table}

\begin{figure}[t]
    \centering
    \includegraphics[width=\linewidth]{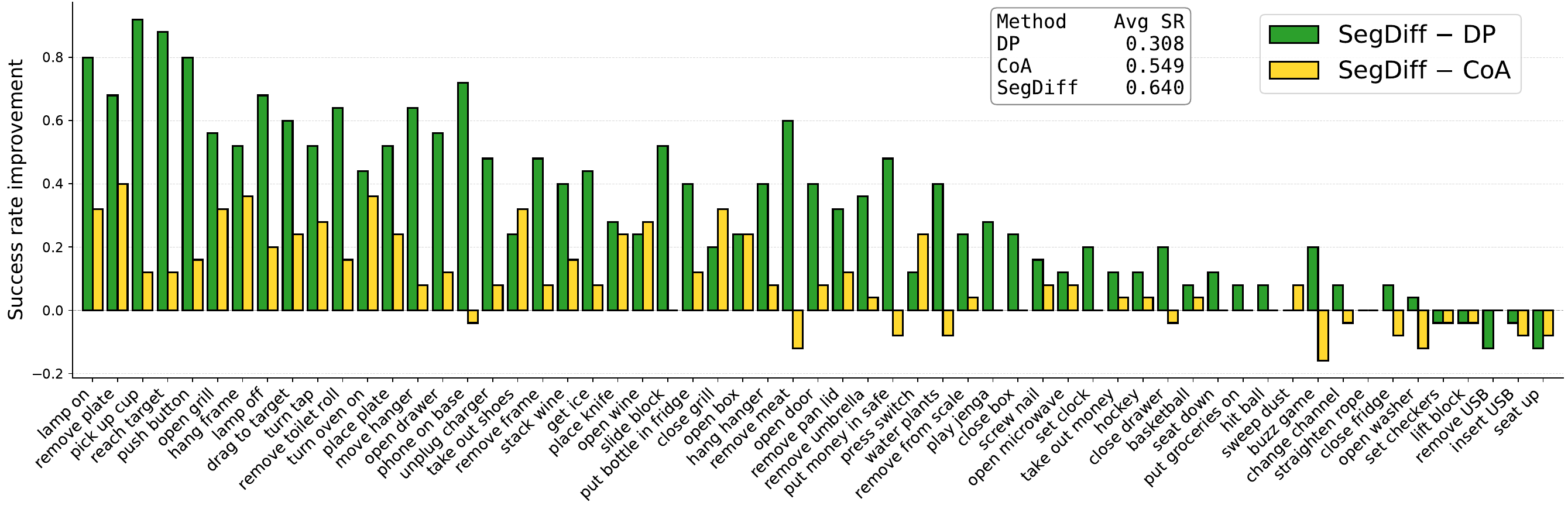}
    \caption{Success rate improvement of SegDiff over DP/CoA. Detailed per-task results are provided in the supplementary material.}
    \label{fig:improve}
\end{figure}

\section{Experiment}
\label{sec:experiment}
\subsection{Simulation Experiments}
\label{sec:sim}

\noindent\textbf{Benchmarks.} We evaluate SegDiff on two simulation benchmarks: RLBench~\cite{james2020rlbench} and RoboMimic~\cite{mandlekar2022matters}. \textbf{RLBench} is built atop CoppeliaSim~\cite{rohmer2013v}, where a Franka Panda robot manipulates the scene. We follow the task setup of CoA~\cite{zhang2025chain} and evaluate on both 10 representative tasks and the full RLBench-60 benchmark. RLBench demonstrations are generated via scripted policies, providing precise and consistent supervision. Each task contains 100 training episodes and 25 evaluation episodes. \textbf{RoboMimic} is another widely used benchmark for evaluating visuomotor policies. We conduct experiments on three tasks using the PH dataset, which consists of human expert demonstrations collected via teleoperation. This allows us to evaluate SegDiff on datasets with different characteristics, ranging from scripted demonstrations to human-collected trajectories, highlighting its robustness across diverse data sources.

\vspace{0.05in}
\noindent\textbf{Baselines.} On RLBench, we evaluate our method against four representative visuomotor policies: ACT~\cite{zhao2023learning}, Diffusion Policy (DP)~\cite{chi2025diffusion}, Chain-of-Action (CoA)~\cite{zhang2025chain}, and a fine-tuned Octo policy~\cite{zhang2025effective}. Among the baselines, ACT and DP are the most commonly used baselines in prior work. CoA and the fine-tuned Octo are among the strongest previously reported methods on this benchmark. 
On RoboMimic, we compare our method against DP and AWE~\cite{shi2023waypoint}, where AWE enhances DP by filtering demonstration trajectories to emphasize critical waypoints. This comparison highlights the effectiveness of Segmented Trajectory Modeling over existing continuous diffusion-based approaches.

\noindent\textbf{Implementation Details.} Following prior work~\cite{chi2025diffusion}, SegDiff utilizes a separate ResNet18 encoder (not pretrained) for each camera view to process historical visual observations. The BatchNorm layers in the original ResNet blocks are replaced by GroupNorm~\cite{wu2018group}, and the final global average pooling is substituted with spatial softmax pooling~\cite{mandlekar2022matters} to better capture spatial information. On RLBench, visual observations are captured by four RGB cameras (front, left shoulder, right shoulder, and wrist) and the images are rendered at a resolution of $128 \times 128$.  On RoboMimic, observations are captured by a third-person camera and a wrist camera with a resolution of $84 \times 84$. Additional details are provided in the supplementary material.

\begin{figure}[b]
    \centering
    \includegraphics[width=\linewidth]{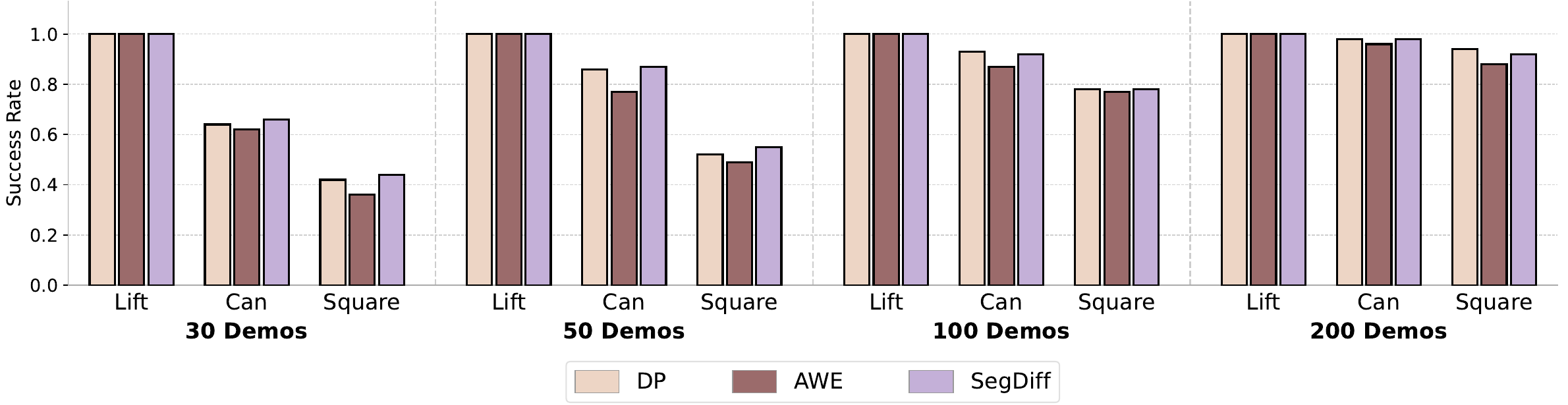}
    \caption{Success rate on RoboMimic with varying numbers of demonstrations.}
    \label{fig:robomimic}
\end{figure}

\vspace{0.05in}
\noindent\textbf{Quantitative Results.} Tab.~\ref{tab:rlbench} demonstrates that SegDiff consistently outperforms all baselines across 10 RLBench tasks, achieving an 11.1\% higher average success rate and the best performance on 9 tasks. Notably, SegDiff yields over 30\% improvement in precision-sensitive tasks (e.g., \textit{Turn Tap}) and outperforms Diffusion Policy (DP) by 30\% on average, validating the efficacy of Segmented Trajectory Modeling for high-precision manipulation. Furthermore, as shown in Fig.~\ref{fig:improve}, SegDiff excels on the full RLBench-60 benchmark, outperforming DP and CoA by 34\% and 9\% respectively, demonstrating improved robustness and generalization.
On RoboMimic (Fig.~\ref{fig:robomimic}), SegDiff exhibits better data efficiency in low-demo regimes by mitigating compounding errors through keypose-anchored trajectory modeling, while remaining highly competitive as demonstrations increase. Detailed training configurations, task descriptions, and exhaustive results are provided in the supplementary material.

\subsection{Real-World Experiments}
\label{sec:real}

\begin{figure}[t]
    \centering
    \includegraphics[width=\linewidth]{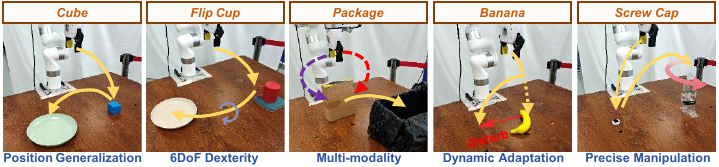}
    \caption{Demonstrations and design of real-world tasks.}
    \label{fig:real}
\end{figure}

\noindent\textbf{Real-World Benchmarks.}
To comprehensively evaluate SegDiff, we design five real-world manipulation tasks (Fig.~\ref{fig:real}):

1) \textbf{\textit{Cube:}} Grasp and place a cube onto a plate. Demonstrations are collected in a fixed area to evaluate robustness to \textbf{position generalization}.

2) \textbf{\textit{Flip Cup:}} Grasp, flip, and place an inverted cup. The randomized handle orientation tests complex \textbf{6-DoF} manipulation.

3) \textbf{\textit{Package:}} Place a cardboard into a box. Two distinct sets of demonstrations (grasping from left/right) were provided to evaluate the policy's ability to handle \textbf{multi-modal} action distributions.

4) \textbf{\textit{Banana:}} Pick up a banana. The pose of the banana is actively perturbed during execution to test \textbf{real-time adaptation to dynamic scenarios}.

5) \textbf{\textit{Screw Cap:}} Grasp and screw a cap onto a bottle. The initial position of the cap is randomized to test proficiency in \textbf{high-precision} operations.

\vspace{0.05in}
\noindent\textbf{Details.} 
Our robotic platform is an xArm7 manipulator equipped with a gripper. Visual inputs are captured by two Intel RealSense D435i cameras (third-person and wrist-mounted), with images collected via human teleoperation and resized to $240 \times 240$. We use the same model architecture as described in Sec.~\ref{sec:sim}.

\vspace{0.05in}
\noindent\textbf{Quantitative Results and Analysis.} Real-world comparative results are presented in Tab.~\ref{tab:real}, where each task is evaluated with 10 trials. We specifically include DP (w/ TE) and DP (w/o TE) to illustrate the effect of \textbf{temporal ensembling (TE)}. Overall, SegDiff achieves performance superior or comparable to all baselines. Task-level analysis follows:

For \textbf{\textit{Cube}} (\textbf{position generalization}), SegDiff achieves a performance comparable to existing approaches.

For \textbf{\textit{Flip Cup}} (\textbf{6-DoF}), SegDiff achieves a higher success rate by accurately planning the motion to the precise grasp pose for the cup handle.

For \textbf{\textit{Package}} (\textbf{multi-modality}), since DP (w/o TE) does not incorporate temporal ensembling, the robot may exhibit an initial period of dithering around the starting position; however, once the policy commits to a specific behavioral mode (\ie, move to the left side), it can typically continue execution until successful completion. DP (w/TE) and ACT often fail due to the oscillation between different action modalities, leading to OOD states. SegDiff with Dynamic Temporal Ensembling prevents modality switching, achieving a success rate of 100\%.

For \textbf{\textit{Banana}} (\textbf{dynamic adaptation}), DP (w/TE) and ACT are negatively influenced by the outdated action buffer. The short prediction horizon of DP (w/o TE) limits its ability to react promptly. SegDiff recovers faster and executes correct actions due to its long-horizon planning and its ability to promptly discard the unusable action buffer.

For \textbf{\textit{Screw Cap}} (\textbf{precise manipulation}), SegDiff is the only method that achieves partial success, indicating its advantage in precise manipulation.

The real-world experiments confirm the effectiveness and manipulation capabilities of SegDiff in the physical world. Additional qualitative visualizations of full task executions are provided in the supplementary material.

\begin{table}[h]

\begin{center}
\setlength{\tabcolsep}{2pt}

\begin{minipage}{0.52\linewidth}
\begin{center}
{
\caption{Success rates on real-world tasks.}

\begin{tabular}{lcccc}
\toprule
Task & \makecell[c]{DP \\ (w/ TE)} & \makecell[c]{DP \\ (w/o TE)} & ACT & SegDiff \\
\midrule
Cube      & 0.7 & 0.6 & 0.7 & \textbf{0.8} \\
Flip Cup  & 0.2 & 0.6 & 0.3 & \textbf{0.9} \\
Package   & 0.4 & 0.8 & 0.8 & \textbf{1.0} \\
Banana    & 0.2 & 0.4 & 0.4 & \textbf{0.9} \\
Screw Cap & 0.0 & 0.0 & 0.0 & \textbf{0.6} \\
\bottomrule
\end{tabular}

\label{tab:real}
}
\end{center}
\end{minipage}
\hfill
\begin{minipage}{0.47\linewidth}
\begin{center}
{
\caption{Ablation on DTE mechanism.}

\begin{tabular}{lccc}
\toprule
Task & RAND & INV & DTE \\
\midrule
RLBench   & \textbf{0.880} & 0.800 & 0.860 \\
Cube      & \textbf{0.8}   & 0.7   & \textbf{0.8} \\
Flip Cup  & \textbf{0.9}   & 0.7   & \textbf{0.9} \\
Package   & 0.8            & \textbf{1.0} & \textbf{1.0} \\
Banana    & 0.5            & 0.3   & \textbf{0.9} \\
Screw Cap & 0.3            & 0.5   & \textbf{0.6} \\
\bottomrule
\end{tabular}

\label{tab:dte}
}
\end{center}
\end{minipage}
\end{center}
\end{table}

\section{Ablation Study and Discussion}
\label{sec:ablation}
\subsection{Segmented Trajectory Modeling}
\label{sec:ablation-stm}

\begin{figure*}[h]
    \centering
    \includegraphics[width=\linewidth]{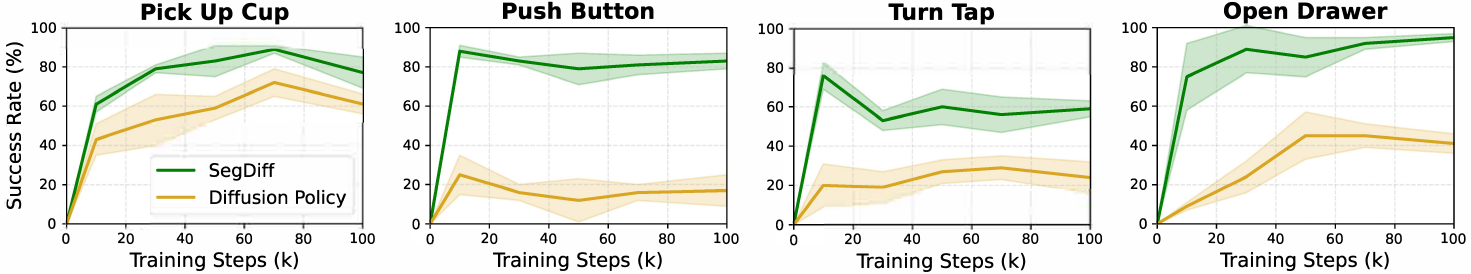}
    \caption{Learning curves of SegDiff and Diffusion Policy across 4 tasks on RLBench.}
    \label{fig:lc}
\end{figure*}

A core component of SegDiff is \textbf{Segmented Trajectory Modeling}. We use Diffusion Policy (DP) as the baseline, which corresponds to a version without Segmented Trajectory Modeling. For fair comparison, both models employ only the basic temporal ensembling, with the Dynamic Temporal Ensembling mechanism disabled. Fig.~\ref{fig:lc} shows learning curves on four selected RLBench tasks, demonstrating that Segmented Trajectory Modeling accelerates convergence and achieves higher performance with fewer training steps.

\subsection{Dynamic Temporal Ensembling}
\label{sec:ablation-dte}

We conduct ablation experiments on the \textbf{Dynamic Temporal Ensembling} (DTE) mechanism by comparing three strategies:

1) \textbf{RAND}: Actions are sampled from random Gaussian noise and executed using vanilla temporal ensembling.

2) \textbf{INV}: Random sampling is used only when no action buffer is available (segment start); afterward, predictions are initialized from the inverted noise.

3) \textbf{DTE}: The complete DTE mechanism is applied.

We use the average success rate on 10 RLBench tasks as the primary metric for simulation. As shown in Tab.~\ref{tab:dte}, the DTE mechanism yields no performance improvement in the simulation environment. This is expected, as simulation environments are generally perturbation-free and exhibit highly monomodal trajectories. In contrast, DTE demonstrates significant advantages in the real-world setting. Specifically, DTE substantially boosts the success rate in the dynamically perturbed \textbf{Banana} task. For the \textbf{Package} task, which features multi-modal trajectories, DDIM inversion allows both the DTE and INV modes to perform markedly better than RAND. The performance difference between DTE and RAND is marginal for the remaining tasks with relatively fixed trajectory patterns.  Crucially, both simulation and real-world results indicate that, in simpler tasks with deterministic trajectories, the INV mode negatively impacts performance. This finding aligns with our theoretical analysis in Sec.~\ref{sec:method-dte}, as the action sampled from inverted noise is potentially influenced by observation from earlier timesteps, consequently reducing the policy's real-time adaptiveness.

\subsection{Hyperparameter Sensitivity}
\label{sec:ablation-hyper}
We also conduct several additional experiments on RLBench to analyze the sensitivity of key hyperparameters. The results are computed using the average success rate across 10 RLBench tasks.

\begin{figure}[t]
    \centering
    \begin{subfigure}[t]{0.48\linewidth}
        \centering
        \includegraphics[width=0.85\linewidth]{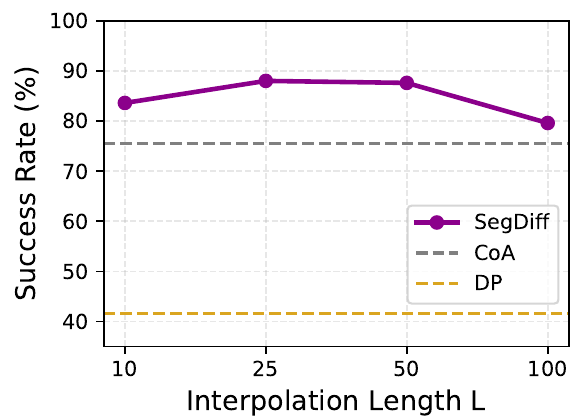}
        \label{fig:ablation_L}
    \end{subfigure}
    \hfill
    \begin{subfigure}[t]{0.48\linewidth}
        \centering
        \includegraphics[width=0.85\linewidth]{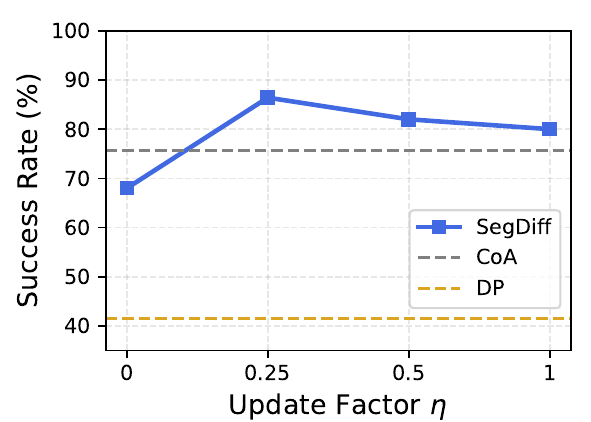}
        \label{fig:ablation_eta}
    \end{subfigure}
    \caption{Results on different interpolation length ($L$) and update factor ($\eta$).}
    \label{fig:ablation}
\end{figure}

For the trajectory interpolation length ($L$) used in Segmented Trajectory Modeling, the left panel of Fig.~\ref{fig:ablation} shows that the success rate peaks at $L=25$. SegDiff consistently outperforms other methods across different values of $L$, demonstrating the robustness of our proposed approach. 

The update factor ($\eta$) is another significant parameter controlling the blending ratio between the action buffer and the new prediction. As shown in the right panel of Fig.~\ref{fig:ablation}, when $\eta=0$, SegDiff executes the entire predicted trajectory from the current state to the keypose at once (\ie, no receding horizon update). Learning this highly constrained, long-horizon trajectory using a standard diffusion model proves challenging, leading to a noticeable drop in success rate. However, because the prediction still focuses on the critical keypose, its performance remains superior to that of the baseline DP. Conversely, $\eta=1$ corresponds to discarding the action buffer entirely and relying solely on the latest prediction for execution. This also results in degraded performance, as it loses the smoothing and consistency benefits of temporal ensembling. The performance degradation observed at both $\eta=0$ and $\eta=1$ further validates the necessity and efficacy of our \textbf{Dynamic Temporal Ensembling} mechanism.

We also evaluate the sensitivity of the Dynamic Temporal Ensembling (DTE) mechanism to the choice of threshold $\delta$ used for judging whether two trajectories belong to the same mode. Specifically, we compute $\delta$ using three different statistics over the training set: the maximum RMSE (max), the 99\% percentile (p99), and the 95\% percentile (p95) of the RMSEs of all training samples. The results, reported in Tab.~\ref{tab:threshold_ablation}, show that DTE is relatively robust to threshold selection.

\begin{table}[h]
\centering

\begin{minipage}{0.38\linewidth}
\begin{center}
\caption{Results of different threshold ($\delta$) for DTE.}
\begin{tabular}{@{}cccc@{}}
\toprule
$\delta$ & max & p99 & p95 \\
\midrule
Success Rate & \textbf{0.860} & 0.836 & 0.852 \\
\bottomrule
\end{tabular}
\label{tab:threshold_ablation}
\end{center}
\end{minipage}
\hfill
\begin{minipage}{0.6\linewidth}
\begin{center}
\small
\caption{Results of DTE under different sampling and inversion steps. SS: Sampling Steps, IS: Inversion Steps, SR: Success Rate, IT: Inference Time.}
\begin{tabular}{lccc}
\toprule
 & SS=50/IS=50 & SS=50/IS=10 & SS=10/IS=10 \\
\midrule
SR & \textbf{0.860} & \textbf{0.860} & 0.808 \\
IT(ms) & 595.74 & 380.21 & \textbf{121.95} \\
\bottomrule
\end{tabular}
\label{tab:dte_steps}
\end{center}
\end{minipage}
\end{table}

\subsection{Efficiency Analysis}
\label{sec:ablation-time}
\begin{table}[h]
\centering
\caption{Efficiency analysis on a real robot system.}
\label{tab:time}
\resizebox{1.0\linewidth}{!}{\begin{tabular}{lccccccc}
\toprule
Method & Interp. (ms) & Inversion (ms) & Sample (ms) & Control (ms) & Others (ms) & Total (ms) & Mem (GB) \\
\midrule
DP       & --   & --    & 121.74 & 407.98 & 92.45 & 622.17 & 1.07\\
SegDiff  & 0.52 & 21.98 & 125.92 & 408.83 & 98.57 & 655.83 & 1.07\\
\bottomrule
\end{tabular}}
\end{table}

As discussed in Sec.~\ref{sec:method-dte}, the number of DDIM inversion steps in DTE is decoupled from the number of sampling steps. Tab.~\ref{tab:dte_steps} reports the success rate and inference time under different inversion and sampling settings on a single RTX A6000 GPU, indicating that the inversion process is flexible and incurs only minor computational overhead.

Furthermore, we evaluate the runtime efficiency of SegDiff on a real robot system, utilizing 10 steps for DDIM inversion and 50 steps for sampling (Tab.~\ref{tab:time}). Compared to the standard Diffusion Policy (DP), SegDiff increases the total latency by approximately 33.7\,ms ($\sim$5.4\%). Specifically, the trajectory interpolation and DDIM inversion together introduce an overhead of 22.5\,ms per control cycle. The remaining latency arises from the data organization and processing required to sample $\mathbf{\tau}_{\text{rand}}$ and $\mathbf{\tau}_{\text{prior}}$ in parallel. These results demonstrate that SegDiff achieves significantly improved robustness with negligible computational cost, maintaining the real-time responsiveness necessary for high-frequency robot deployment.

\section{Conclusion}

In this paper, we presented SegDiff, a novel framework for robotic manipulation that bridges continuous action prediction and keypose-level constraints via Segmented Trajectory Modeling. By predicting trajectories anchored to the next keypose, SegDiff effectively guides learning toward task-critical regions and alleviates compounding errors. Furthermore, we introduced the Dynamic Temporal Ensembling mechanism, which leverages DDIM inversion to validate and refine the action buffer in real time, ensuring action mode consistency and enhancing responsiveness to dynamic environments. Extensive experiments across diverse manipulation tasks demonstrate the superior performance of SegDiff in both simulation and real-world scenarios. These results highlight the effectiveness of integrating keypose-aware trajectory modeling with dynamic temporal ensembling for generalizable and reliable robotic policy learning. The primary limitation stems from the dependency on pre-defined keyposes for segmenting demonstrations; incorporating automated or self-supervised keypose discovery would therefore be a promising direction for future research.

\clearpage

\bibliographystyle{plainnat}
\bibliography{./main}

\clearpage

\newpage

\beginappendix


\section{Theoretical Analysis of Compounding Errors}
\label{sec:appendix-error-analysis}

As introduced in the method section of the main paper, Segmented Trajectory Modeling (STM) mitigates compounding errors that arise in continuous sliding-window prediction paradigms. In this appendix, we provide a theoretical analysis from the perspective of dynamical systems and imitation learning to support this claim.

\vspace{0.05in}
\noindent\textbf{Problem Setup.}
Consider a deterministic dynamical system governed by

\begin{equation}
s_{t+1} = f(s_t, a_t),
\end{equation}

where $s_t$ denotes the system state and $a_t$ denotes the control action. Let $\pi^*$ be the expert policy and $\pi_\theta$ be the learned policy. For simplicity, we analyze the expected policy output under the diffusion model as a deterministic policy. We assume the learned policy approximates the expert policy with a uniformly bounded error over the state space:

\begin{equation}
\|\pi_\theta(s_t) - \pi^*(s_t)\| \le \epsilon .
\end{equation}

We further assume that the system dynamics are Lipschitz continuous with respect to both state and action:

\begin{equation}
\|f(s,a) - f(s',a')\|
\le
L_s \|s-s'\| + L_a \|a-a'\|.
\end{equation}

Additionally, we assume the expert policy is Lipschitz continuous:

\begin{equation}
\|\pi^*(s) - \pi^*(s')\|
\le
L_\pi \|s-s'\|.
\end{equation}

Let $s_t^*$ denote the expert state trajectory generated by $\pi^*$ and define the state deviation $e_t = \|s_t - s_t^*\|$. Our goal is to analyze how this deviation evolves during closed-loop execution.

\vspace{0.05in}
\noindent\textbf{Error Accumulation in Sliding-Window Paradigms.}
In traditional sliding-window approaches (e.g., Diffusion Policy), the model repeatedly predicts short-horizon action sequences while the prediction window continuously shifts. Because the target trajectory is not anchored to any fixed goal state, small prediction errors can accumulate over time.

The state deviation at timestep $t+1$ satisfies

\begin{equation}
e_{t+1}
=
\| f(s_t, \pi_\theta(s_t)) - f(s_t^*, \pi^*(s_t^*)) \|.
\end{equation}

Using the triangle inequality:

\begin{equation}
\begin{aligned}
e_{t+1}
&\le
\| f(s_t, \pi_\theta(s_t)) - f(s_t, \pi^*(s_t)) \| +
\| f(s_t, \pi^*(s_t)) - f(s_t^*, \pi^*(s_t^*)) \|.
\end{aligned}
\end{equation}

The first term captures the policy prediction error. Using the Lipschitz property of the dynamics with respect to action:

\begin{equation}
\| f(s_t, \pi_\theta(s_t)) - f(s_t, \pi^*(s_t)) \|
\le
L_a \|\pi_\theta(s_t) - \pi^*(s_t)\|
\le
L_a \epsilon .
\end{equation}

The second term captures the propagation of state deviation through the dynamics. Using the Lipschitz property of $f$ and $\pi^*$:

\begin{equation}
\begin{aligned}
\| f(s_t, \pi^*(s_t)) - f(s_t^*, \pi^*(s_t^*)) \| &\le
L_s \|s_t - s_t^*\|
+
L_a \|\pi^*(s_t) - \pi^*(s_t^*)\| \\
&\le
L_s e_t + L_a L_\pi e_t .
\end{aligned}
\end{equation}

Combining the two bounds gives

\begin{equation}
e_{t+1}
\le
(L_s + L_a L_\pi) e_t + L_a \epsilon .
\end{equation}

Define the closed-loop Lipschitz constant

\begin{equation}
\tilde L = L_s + L_a L_\pi .
\end{equation}

The error recursion becomes

\begin{equation}
e_{t+1} \le \tilde L e_t + L_a \epsilon .
\end{equation}

Solving this recurrence relation with $e_0 = 0$ yields

\begin{equation}
e_t
\le
L_a \epsilon
\sum_{i=0}^{t-1} \tilde L^i .
\end{equation}

For $\tilde L \neq 1$, this geometric series evaluates to:
\begin{equation}
e_t \le L_a \epsilon \frac{\tilde L^t - 1}{\tilde L - 1}.
\end{equation}

If $\tilde L > 1$, the state deviation $e_t$ grows exponentially with respect to time $t$, making long-horizon open-loop predictions extremely brittle. Even in the marginally stable case ($\tilde L = 1$), which is a standard assumption in kinematic control analyses, the summation simplifies to a linear growth:
\begin{equation}
e_t \le t L_a \epsilon.
\end{equation}

Under this $\tilde L = 1$ assumption, the cumulative trajectory deviation over the entire execution horizon $T$ becomes:
\begin{equation}
\mathcal{E}_{\text{sliding}} = \sum_{t=1}^{T} e_t \le \sum_{t=1}^{T} t L_a \epsilon = \frac{T(T+1)}{2} L_a \epsilon = \mathcal{O}(L_a \epsilon T^2).
\end{equation}

This at least quadratic growth behavior (which becomes exponential if $\tilde L > 1$) formally illustrates the well-known compounding error phenomenon in behavior cloning: small per-step prediction errors accumulate rapidly over long horizons, eventually driving the system into out-of-distribution (OOD) states.

\vspace{0.05in}
\noindent\textbf{Error Mitigation via Segmented Trajectory Modeling (STM).}
To address the compounding errors of sliding-window approaches, SegDiff decomposes the full trajectory of length $T$ into $M$ segments separated by keyposes $k_0, k_1, \dots, k_M$, where $k_0 = 0$ and $k_M = T$. Let the maximum length of any single segment be bounded by $S$, such that $k_m - k_{m-1} \le S < T$.

Consider the system at timestep $t$ within the $m$-th segment, i.e., $t \in [k_{m-1}, k_m)$. STM predicts the action sequence $\hat{a}_{t:k_m-1}$ anchored at the next keypose $k_m$. To formally derive the bound on the keypose prediction error $\Delta_{k_m|t} = \|\hat{s}_{k_m|t} - s_{k_m}^*\|$, we trace the intermediate predicted states $\hat{s}_{\tau|t}$ for $\tau \in \{t, \dots, k_m\}$ starting from $\hat{s}_{t|t} = s_t$. The deviation satisfies the following inequality:

\begin{equation}
\begin{aligned}
\|\hat{s}_{\tau+1|t} - s_{\tau+1}^*\|
&= \|f(\hat{s}_{\tau|t}, \hat{a}_\tau) - f(s_\tau^*, a_\tau^*)\| \\
&\le L_s \|\hat{s}_{\tau|t} - s_\tau^*\| + L_a \|\hat{a}_\tau - a_\tau^*\| \\
&\le L_s \|\hat{s}_{\tau|t} - s_\tau^*\| + L_a \epsilon .
\end{aligned}
\end{equation}

Applying the inequality recursively from $t$ to $k_m - 1$ yields, the keypose prediction error is bounded by

\begin{equation}
\Delta_{k_m|t}
\le
L_s^{k_m-t} e_t
+
L_a \epsilon
\sum_{i=0}^{k_m-t-1} L_s^i .
\label{eq:delta_km_t}
\end{equation}

Note that within each segment, the error propagation is governed only by the open-loop dynamics constant $L_s$, whereas in sliding-window paradigms it is governed by the closed-loop constant $\tilde{L} = L_s + L_a L_\pi$. Since $\tilde{L} > L_s$, SegDiff inherently reduces the rate of error divergence.

Furthermore, keyposes in robotic manipulation often act as bottleneck states that physically absorb relative pose errors, effectively resetting the execution deviation at the beginning of each segment ($e_{k_{m-1}} \approx 0$). Under this assumption, the deviation $e_t$ within the segment satisfies

\begin{equation}
e_t
\le
L_a \epsilon
\sum_{j=0}^{t-k_{m-1}-1} L_s^j .
\end{equation}

Substituting this bound into Eq.~\ref{eq:delta_km_t} yields

\begin{equation}
\begin{aligned}
\Delta_{k_m|t}
&\le
L_s^{k_m-t}
\left(
L_a \epsilon
\sum_{j=0}^{t-k_{m-1}-1} L_s^j
\right)
+
L_a \epsilon
\sum_{i=0}^{k_m-t-1} L_s^i \\
&=
L_a \epsilon
\left(
\sum_{j=0}^{t-k_{m-1}-1} L_s^{k_m-t+j}
+
\sum_{i=0}^{k_m-t-1} L_s^i
\right) \\
&=
L_a \epsilon
\sum_{n=0}^{k_m-k_{m-1}-1} L_s^n
\le
L_a \epsilon
\sum_{n=0}^{S-1} L_s^n .
\end{aligned}
\end{equation}

The prediction error of the next keypose $\Delta_{k_m|t}$ is strictly bounded by the local segment length $S$. In the marginally stable case ($L_s = 1$), this simplifies to $\Delta_{k_m|t} \le S L_a \epsilon$. Even in unstable regimes where $L_s > 1$, STM still provides a significantly tighter bound than sliding-window paradigms. Recall that sliding-window methods suffer from error growth governed by the closed-loop constant $\tilde{L} = L_s + L_a L_\pi$. Since $\tilde{L} > L_s$, the exponential divergence in SegDiff is restricted to the open-loop dynamics within a short segment $S$, avoiding the accelerated drift caused by the learned policy's sensitivity $L_\pi$ over a long horizon.

Unlike sliding-window paradigms where errors propagate continuously and influence all future states, STM decomposes the trajectory into nearly independent segments through keypose anchoring. In manipulation tasks, the successful execution of each segment primarily depends on whether the robot reaches the target keypose that terminates the segment. Therefore, the execution quality within a segment can be characterized by the accuracy of the predicted next keypose.

During execution, the policy predicts the future trajectory toward the same keypose multiple times at different timesteps within the segment. SegDiff further aggregates these repeated predictions through the Dynamic Temporal Ensembling (DTE) mechanism, producing a temporally smoothed estimate of the next keypose. Since the final prediction is obtained by averaging multiple predictions generated within the same segment, its deviation is naturally bounded by the largest single prediction error among them. Consequently, the execution stability of a segment is determined by the worst-case keypose prediction error that occurs within that segment. From this perspective, the global execution error of the entire trajectory can be bounded by

\begin{equation}
\mathcal{E}_{\text{SegDiff}}
\leq
\max_{m \in \{1,\dots,M\}}
\left(
\max_{t \in [k_{m-1}, k_m-1]}
\Delta_{k_m|t}
\right)
\le
\mathcal{O}(L_a \epsilon S).
\end{equation}

In contrast to sliding-window paradigms where the cumulative deviation scales with the full execution horizon $T$ (i.e., $\mathcal{O}(L_a \epsilon T^2)$ under marginal stability), the error in STM is strictly capped by the local segment scale $S$. Since $S \ll T$ in long-horizon manipulation tasks, SegDiff effectively transforms a long-term compounding error problem into a sequence of short-horizon, locally bounded regulation problems, thereby substantially improving robustness during long-horizon execution.

\section{Additional Visualizations}

\subsection{Visualization of Dynamic Temporal Ensembling Mechanism}
As discussed in the method section, the Dynamic Temporal Ensembling (DTE) mechanism is designed to address potential action mode conflicts and enhance adaptation to dynamic environments.

\begin{figure}[h]
    \centering
    
    \begin{subfigure}{0.33\linewidth}
        \centering
        \includegraphics[width=\linewidth]{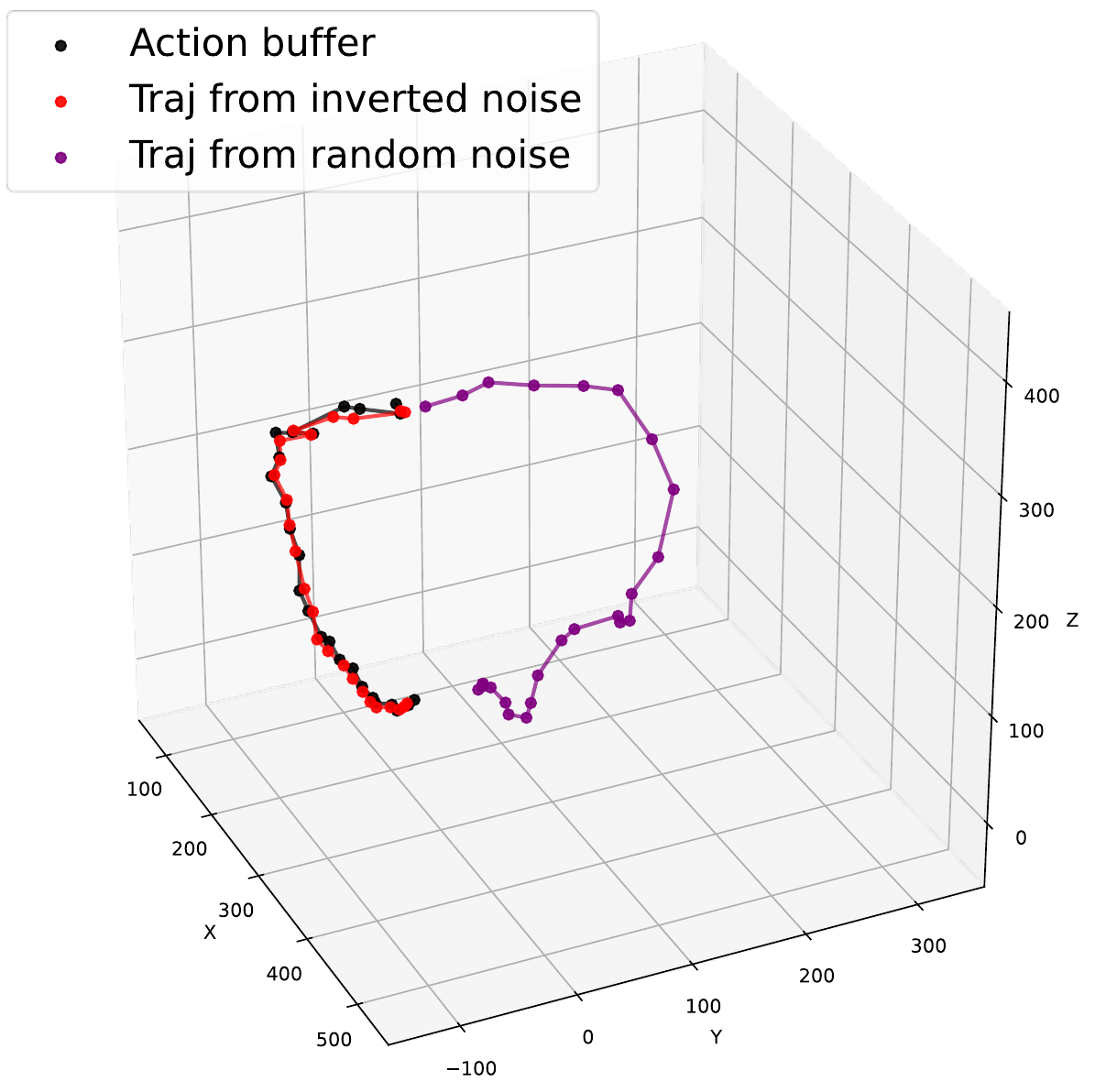}
        \caption{Sampling from random Gaussian noise may generate a correct trajectory but inconsistent with the action buffer.}
        \label{fig:demo_mode}
    \end{subfigure}
    \hfill
    \begin{subfigure}{0.66\linewidth}
        \centering
        \includegraphics[width=\linewidth]{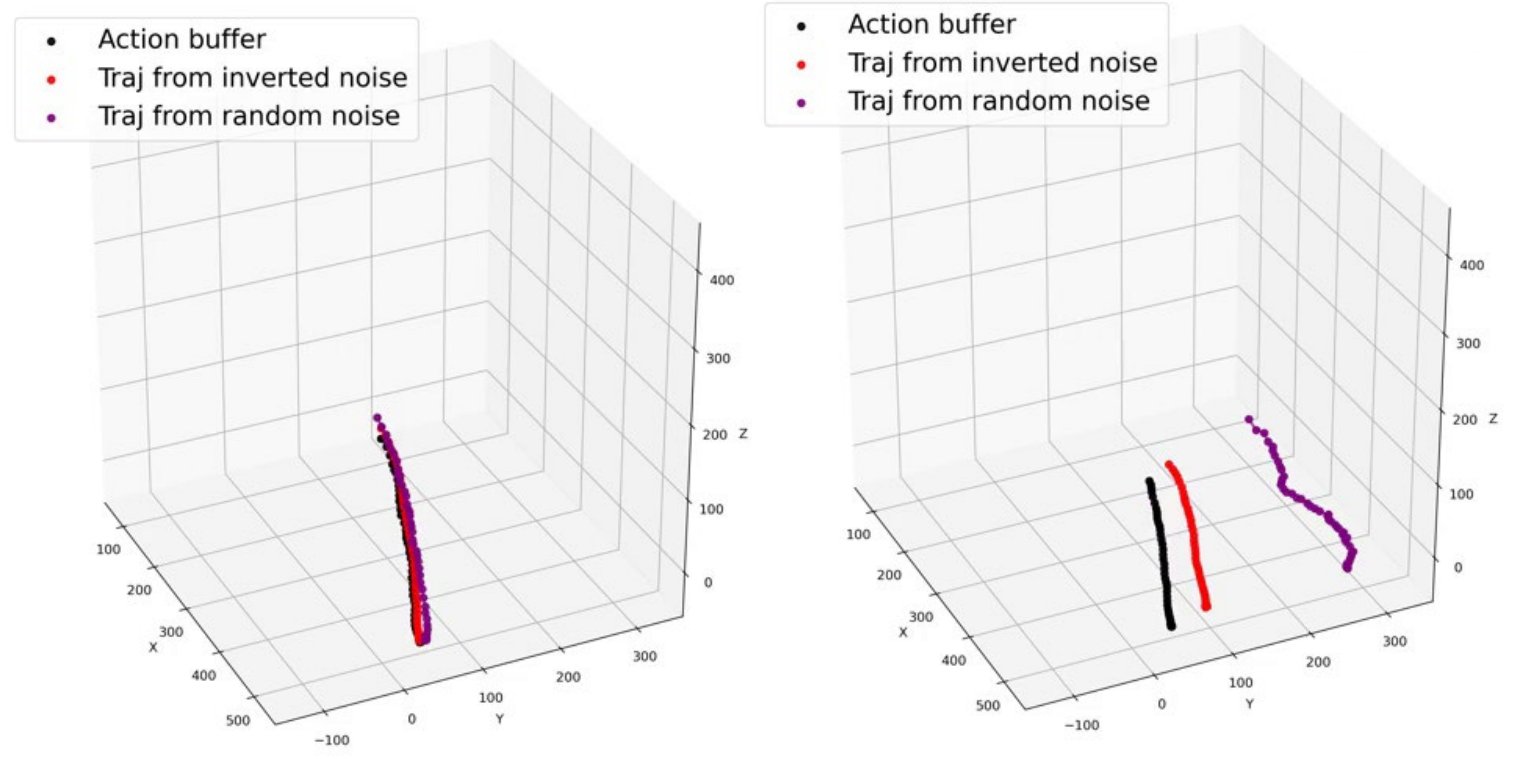}
        \caption{Two situations faced by Dynamic Temporal Ensembling (left: valid action buffer; right: invalid buffer that must be discarded).}
        \label{fig:te_demo}
    \end{subfigure}
    
    \caption{Illustrations of issues encountered during inference with diffusion-based policies and how Dynamic Temporal Ensembling handles them.}
    \label{fig:te_cases}
\end{figure}

In tasks such as \textbf{Package} mentioned in the experiment section, the same observation may correspond to multiple feasible action modes—for example, placing the package from different approach directions. When trajectories are sampled purely from random Gaussian noise, the resulting action may follow a mode that differs from the one encoded in the current action buffer (as illustrated in Fig.~\ref{fig:demo_mode}). Interpolating between two inconsistent modes can lead to undesirable behavior. To prevent this, DTE initializes the sampling process using the noise inverted from the buffered trajectory, ensuring that the newly generated trajectory remains consistent with the buffer, as long as the buffer is still valid under the latest observation.

The left panel of Fig.~\ref{fig:te_demo} shows the situation where the action buffer is valid, and the trajectory sampled from the inverted noise is aligned with the action mode of the existing buffer. In such cases, the action buffer is refined through temporal ensembling, resulting in smoother and more accurate action output. The right panel illustrates the situation when the environment is significantly altered (generated in the execution process of the \textbf{Banana} task). In such cases, the action buffer is discarded, allowing the policy to rely solely on predictions inferred from the latest observation.

\subsection{Visualization of SegDiff Execution} 
\begin{figure}[h]
    \centering
    \includegraphics[width=\linewidth]{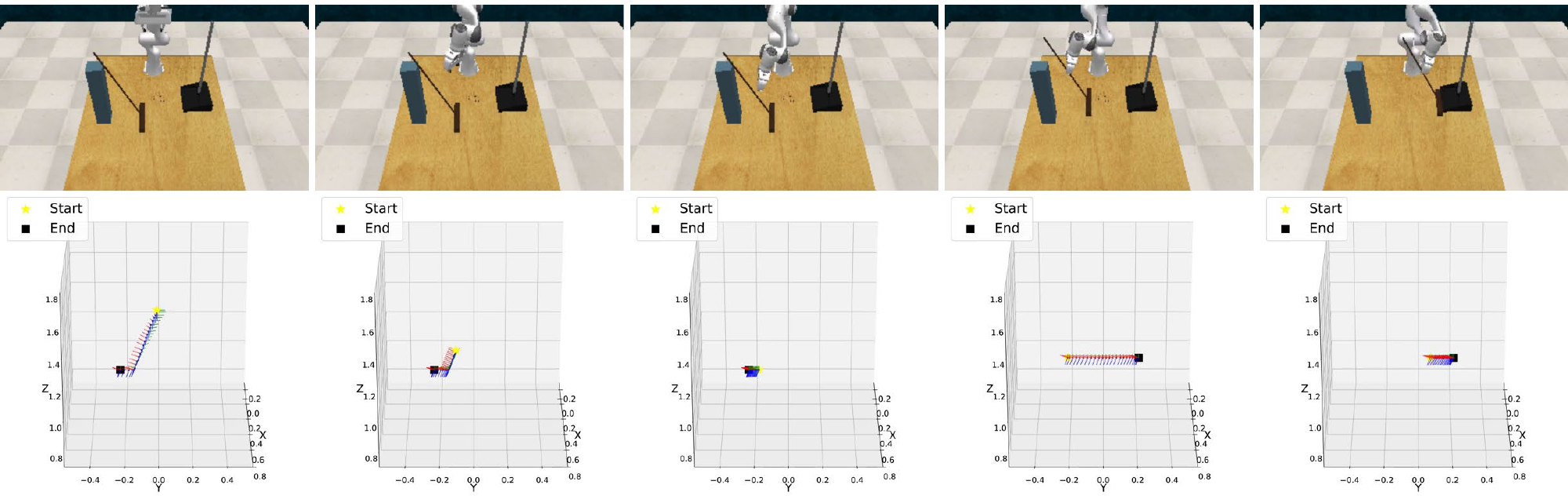}
    \caption{Visualization of the predicted pose trajectory in Segmented Trajectory Modeling (STM). The prediction always targets at the next keypose, ensuring focused modeling of the trajectory toward task-critical points.}
    \label{fig:exec}
\end{figure}

As detailed in the method section, SegDiff utilizes Segmented Trajectory Modeling (STM), predicting the continuous trajectory from the current state to the next keypose. This approach allows the model to focus on critical points while simultaneously modeling the intermediate motion. The visualization of the SegDiff execution process, showing the predicted segmented trajectory and corresponding observation, is provided in Fig.~\ref{fig:exec}.

\section{Task Details}

\subsection{RLBench Tasks}

The full task names, descriptions, and number of corresponding keyposes of the selected 10 RLBench tasks are as follows:

\noindent (1) \textbf{Sweep Dust} (sweep to dustpan): use the broom to brush the dirt into the dustpan. This task involves 2 to 3 keyposes.

\noindent (2) \textbf{Push Button} (push button): press the button with the maroon base. This task involves 2 keyposes.

\noindent (3) \textbf{Stack Wine} (stack wine): slide the bottle onto the wine rack. This task involves 2 keyposes.

\noindent (4) \textbf{Turn Tap} (turn tap): grasp the tap and turn it. This task involves 2 keyposes.

\noindent (5) \textbf{Open Drawer} (open drawer): grip the bottom handle and pull the drawer open. This task involves 2 keyposes.

\noindent (6) \textbf{Pick Up Cup} (pick up cup): grasp the red cup and lift it. This task involves 2 keyposes.

\noindent (7) \textbf{Take Lid} (take lid off saucepan): using the handle, lift the lid off of the pan. This task involves 2 keyposes.

\noindent (8) \textbf{Press Switch} (press switch): turn the switch on or off. This task involves 2 keyposes.

\noindent (9) \textbf{Reach Target} (reach target): touch the red ball with the panda gripper. This task involves only 1 keypose.

\noindent (10) \textbf{Open Box} (open box): open the box lid. This task involves 2 keyposes.

\subsection{RoboMimic Tasks}

 Descriptions and number of corresponding keyposes of three tasks from RoboMimic are as follows:

\noindent (1) \textbf{Lift}: grasp the cube from the table. This task involves 2 keyposes.

\noindent (2) \textbf{Can}: grasp the can and then place it in the designated place. This task involves 3 keyposes.

\noindent (3) \textbf{Square}: pick up the square and assemble it with the wooden peg. This task involves 3 keyposes.

\begin{figure}[h]
    \centering
    \includegraphics[width=\linewidth]{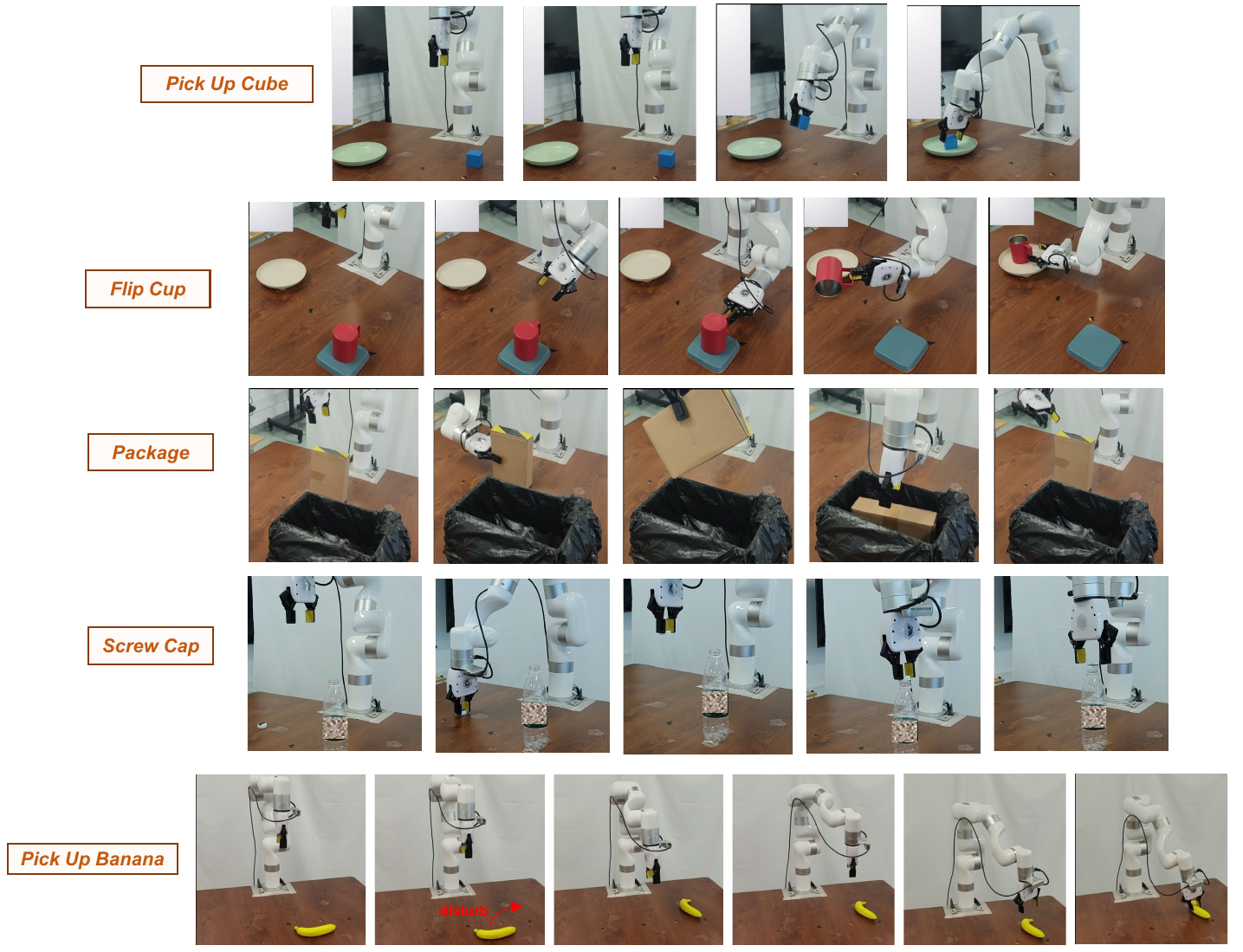}
    \caption{Visualization of the complete processes of real-world tasks.}
    \label{fig:real_task}
\end{figure}

\subsection{Real-World Tasks}

The visualization of the process of all real-world tasks can be found in Fig.~\ref{fig:real_task}, and the detailed descriptions are as follows:

\noindent (1) \textbf{Cube:} Grasping and placing a cube onto a plate. Demonstrations were collected in a specific area to test the ability to handle \textbf{position changes} in the workspace.

\noindent (2) \textbf{Flip Cup:} Grasping an inverted cup, flipping it mid-air, and placing it. The initial handle orientation was randomized to test complex \textbf{6-DoF} manipulation.

\noindent (3) \textbf{Package:} Placing a cardboard into a box. Two distinct sets of demonstrations (10 samples each) were provided to evaluate the policy's ability to handle \textbf{multi-modality} in action distributions.

\noindent (4) \textbf{Banana:} Picking up a banana. The pose of the banana is actively perturbed during execution to test \textbf{real-time adaptation to dynamic scenarios}.

\noindent (5) \textbf{Screw Cap:} Grasping a cap and screwing it onto a bottle. The initial position of the cap is randomized to test proficiency in \textbf{high-precision} operations.

\section{Hyperparameters and Experiment Settings}
\noindent\textbf{RLBench} We provide the training and evaluation hyperparameters for SegDiff and DP used on RLBench. The hyperparameters are aligned with previous works~\cite{zhang2025chain, shridhar2025generative, chi2025diffusion}, allowing us to isolate and assess the impact of our proposed Segmented Trajectory Modeling. Specifically, SegDiff inherits most of its settings from DP. Additionally, we incorporate vanilla temporal ensembling into DP to ensure fair comparison.

\vspace{0.05in}
\noindent\textbf{RoboMimic} We evaluate SegDiff against DP and AWE on the selected three RoboMimic tasks. The hyperparameters are shown in Tab.~\ref{tab:RoboMimic_parameters}.

\begin{table*}[h]
\centering
\scriptsize

\begin{minipage}[t]{0.48\textwidth}
\centering
\caption{Hyperparameters for SegDiff and DP on RLBench}
\label{tab:RLBench_parameters}
\resizebox{\linewidth}{!}{%
\begin{tabular}{@{}lcc@{}}
\toprule
\textbf{Parameter} & \textbf{SegDiff} & \textbf{DP} \\
\midrule
Action dimension & 8 & 8 \\
Cameras & 4 & 4 \\
Image size & $128 \times 128$ & $128 \times 128$ \\
Observation horizon & 1 & 1 \\
Action sequence length & \text{---} & 20 \\
Interpolation length ($L$) & 25 & \text{---} \\
Batch size & 128 & 128 \\
Learning rate & $1e^{-4}$ & $1e^{-4}$ \\
Weight decay & $1e^{-4}$ & $1e^{-4}$ \\ 
Iteration & 100000 & 100000 \\ 
Hidden dim & 256 & 256 \\
Feedforward dim & 1024 & 1024 \\
Noise predictor & Transformer & Transformer \\
Scheduler & DDPM & DDPM \\
Train diffusion steps & 50 & 50 \\
\midrule
\multicolumn{3}{l}{\textbf{Inference\&Execution Parameters}} \\
Num. execution ($n$) & 1 & 1 \\
Update factor ($\eta$) & 0.25 & \text{---} \\
Threshold ($\delta$) & max & \text{---} \\
Sampling steps ($SS$) & 50 & 50 \\
Inversion steps ($IS$) & 10 & \text{---} \\
\bottomrule
\end{tabular}%
}
\end{minipage}
\hfill
\begin{minipage}[t]{0.48\textwidth}
\centering
\caption{Hyperparameters for SegDiff, DP, and AWE on RoboMimic}
\label{tab:RoboMimic_parameters}
\resizebox{\linewidth}{!}{%
\begin{tabular}{@{}lccc@{}}
\toprule
\textbf{Parameter} & \textbf{SegDiff} & \textbf{DP} & \textbf{AWE} \\
\midrule
Action dimension & 8 & 8 & 8 \\
Cameras & 2 & 2 & 2 \\
Image size & $76 \times 76$ & $76 \times 76$ & $76 \times 76$ \\
Observation history & 2 & 2 & 2 \\
Horizon & \text{---} & 10 & 10 \\
Action sequence length & \text{---} & 8 & 8 \\
Interpolation length & 50 & \text{---} & \text{---} \\
Batch size & 64 & 64 & 64 \\
Learning rate & $1.0e^{-4}$ & $1.0e^{-4}$ & $1.0e^{-4}$ \\
Weight decay & $1e^{-3}$ & $1e^{-3}$ & $1e^{-3}$ \\ 
Epoch & 2000 & 2000 & 2000 \\
Hidden dim & 256 & 256 & 256 \\
Feedforward dim & 1024 & 1024 & 1024 \\
Noise predictor & Transformer & Transformer & Transformer \\
Noise scheduler & DDPM & DDPM & DDPM \\
Train diffusion steps & 100 & 100 & 100 \\
\midrule
\multicolumn{4}{l}{\textbf{Inference \& Execution Parameters}} \\
Num. execution ($n$) & 1 & \text{---} & \text{---} \\
Update factor ($\eta$) & 0.25 & \text{---} & \text{---} \\ 
Threshold ($\delta$) & max & \text{---} & \text{---} \\ 
Sampling steps ($SS$) & 100 & 100 & 100 \\
Inversion steps ($IS$) & 10 & \text{---} & \text{---} \\ 
\bottomrule
\end{tabular}%
}
\end{minipage}

\end{table*}

\begin{table}[t]
\centering
\caption{Statistics of reconstruction distance distribution on RLBench. $\max$ denotes the maximum reconstruction distance; $p99$ and $p95$ denote the 99th and 95th percentiles, respectively.}
\label{tab:threshold_stats}
\setlength{\tabcolsep}{4pt}
\begin{tabular}{@{}lccccccccccc@{}}
\toprule
& \makecell{Sweep \\ Dust} & \makecell{Push \\ Button} & \makecell{Stack \\ Wine} & \makecell{Turn \\ Tap} & \makecell{Open \\ Drawer} & \makecell{Pick Up \\ Cup} & \makecell{Take \\ Lid} & \makecell{Press \\ Switch} & \makecell{Reach \\ Target} & \makecell{Open \\ Box} \\
\midrule
$\max$ & 0.025 & 0.109 & 0.099 & 0.125 & 0.121 & 0.036 & 0.063 & 0.148 & 0.034 & 0.024 \\
${p}99$ & 0.016 & 0.058 & 0.049 & 0.072 & 0.039 & 0.024 & 0.029 & 0.107 & 0.028 & 0.017 \\
${p}95$ & 0.012 & 0.037 & 0.028 & 0.048 & 0.024 & 0.017 & 0.015 & 0.085 & 0.023 & 0.014 \\
\bottomrule
\end{tabular}
\end{table}
\vspace{-10pt}

\section{Additional Experiment Results}
\subsection{DTE Threshold Analysis}
\label{sec:threshold_analysis}
A significant component of our Dynamic Temporal Ensembling (DTE) mechanism is the threshold $\delta$, which is employed to determine whether two trajectories still belong to the same behavioral mode. This determination is crucial for assessing the validity of the actions within the buffer.

To estimate an effective value for $\delta$, we operate on the fundamental assumption that all original trajectories in the training dataset represent correct behaviors conditioned on the given observation. Consequently, the reconstructed trajectory, obtained via DDIM inversion and subsequent sampling, should also belong to the same mode as the original. We statistically estimate the distribution of distances between these reconstructed and original trajectories to establish a quantitative baseline for "same-mode" proximity. While this estimation method based on reconstruction distance is inherently an approximation of the true mode distribution, it provides a measurable proxy.

The DTE mechanism exhibits robustness to the selection of the threshold $\delta$; detailed ablation results in \textbf{Tab.\ref{tab:full}} confirm that the specific choice of $\delta$ does not significantly impact the final task success rate on RLBench.

\begin{figure}[h]
    \centering
    \includegraphics[width=\linewidth]{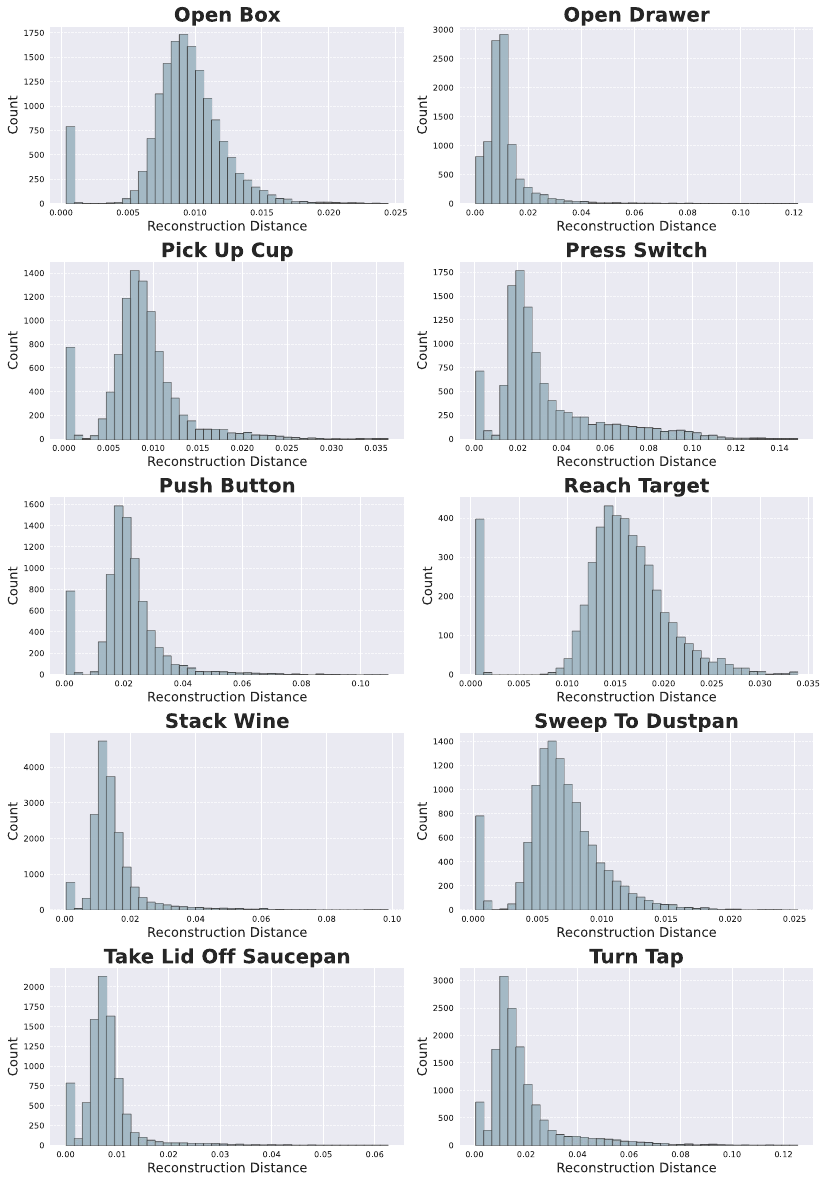}
    \caption{Distribution of reconstruction distance on 10 RLBench tasks.}
    \label{fig:dist}
\end{figure}

The results calculated from the reconstruction distance distribution are shown in Tab.\ref{tab:threshold_stats} and the visualization of the reconstruction distance distribution is shown in Fig.\ref{fig:dist}

\begin{table}[h]
  \caption{Detailed experiment results on RLBench.}
  \setlength{\tabcolsep}{5pt}
  \label{tab:full}
  \centering
  \resizebox{\textwidth}{!}{
  \begin{tabular}{@{}lccccccccccc@{}}
    \toprule
    Method &
    \makecell{Sweep \\ Dust} &
    \makecell{Push \\ Button} &
    \makecell{Stack \\ Wine} &
    \makecell{Turn \\ Tap} &
    \makecell{Open \\ Drawer} &
    \makecell{Pick Up \\ Cup} &
    \makecell{Take \\ Lid} &
    \makecell{Press \\ Switch} &
    \makecell{Reach \\ Target} &
    \makecell{Open \\ Box} &
    Avg. \\
    \midrule
    $\text{SegDiff}$ & 0.96 & 0.76 & 0.88 & 0.72 & 1.00 & 0.96 & 0.80 & 0.80 & 1.00 & 0.72 & 0.860 \\
    $\mathbf{RAND}$ & 1.00 & 0.92 & 0.96 & 0.84 & 1.00 & 0.92 & 0.92 & 0.72 & 0.96 & 0.56 & 0.880 \\
    $\mathbf{INV}$ & 1.00 & 0.88 & 0.80 & 0.60 & 0.88 & 0.60 & 0.88 & 0.76 & 0.96 & 0.64 & 0.800 \\
    $L{=}10$ (w/o DTE) & 0.96 & 0.72 & 0.96 & 0.72 & 0.96 & 0.84 & 0.76 & 0.84 & 1.00 & 0.60 & 0.836 \\
    $L{=}50$ (w/o DTE) & 1.00 & 0.88 & 0.92 & 0.72 & 0.92 & 0.88 & 0.88 & 0.76 & 0.96 & 0.84 & 0.876 \\
    $L{=}100$ (w/o DTE) & 1.00 & 0.76 & 0.96 & 0.60 & 0.80 & 0.92 & 0.88 & 0.52 & 0.96 & 0.56 & 0.796 \\
    $\delta{=}2\times \max$ & 0.96 & 0.84 & 0.88 & 0.68 & 1.00 & 0.96 & 0.80 & 0.80 & 0.96 & 0.60 & 0.848 \\
    $\delta{=}\mathrm{p99}$ & 0.92 & 0.76 & 0.88 & 0.72 & 0.96 & 0.84 & 0.80 & 0.72 & 1.00 & 0.76 & 0.836 \\
    $\delta{=}\mathrm{p95}$ & 0.92 & 0.80 & 0.92 & 0.76 & 0.72 & 0.96 & 0.80 & 0.76 & 1.00 & 0.88 & 0.852 \\
    $50IS/50SS$ & 0.96 & 0.76 & 0.88 & 0.64 & 1.00 & 0.96 & 0.84 & 0.84 & 1.00 & 0.72 & 0.860 \\
    $10IS/10SS$ & 0.96 & 0.76 & 0.92 & 0.48 & 0.92 & 0.84 & 0.84 & 0.36 & 1.00 & 0.56 & 0.764 \\
    $\eta{=}0$ & 0.28 & 0.84 & 0.76 & 0.56 & 0.64 & 0.84 & 0.68 & 0.72 & 1.00 & 0.48 & 0.680 \\
    $\eta{=}0.5$ & 1.00 & 0.88 & 0.80 & 0.56 & 0.96 & 0.76 & 0.88 & 0.84 & 0.96 & 0.60 & 0.824 \\
    $\eta{=}1$ & 0.84 & 0.76 & 0.76 & 0.60 & 1.00 & 0.76 & 0.84 & 0.88 & 0.96 & 0.60 & 0.800 \\
    $\mathbf{linear\ interp}$ & 1.00 & 0.88 & 0.96 & 0.76 & 1.00 & 0.92 & 0.80 & 0.68 & 0.96 & 0.60 & 0.856 \\
    \bottomrule
  \end{tabular}}
\end{table}

\subsection{Detailed Ablation Results}
\label{sec:all_results}
We also study the effect of different interpolation methods. For rotation, we use spherical linear interpolation. For translation, we test both linear and cubic interpolation during training and evaluation. The results, together with those from ablation experiments on 10 RLBench tasks, are shown in Tab.~\ref{tab:full}.
We test different Interpolation Lengths $\mathbf{L}=\{10, 25, 100\}$ without Dynamic Temporal Ensembling (w/o DTE). $\mathbf{RAND}$ and $\mathbf{INV}$ are defined in the experiment section to test the effect of introducing DDIM inversion. We also test the mode switch threshold $\mathbf{e}=\{2*\max, \mathrm{p99}, \mathrm{p95}\}$, and the update factor $\mathbf{\eta}=\{0, 0.5, 1\}$ used for blending trajectories. Additionally, we test the efficiency trade-off using different numbers of inversion steps ($\mathbf{IS}$) and sampling steps ($\mathbf{SS}$). Finally, $\mathbf{linear\ interp}$ uses linear interpolation for translation compared to the default cubic interpolation.

\clearpage
\section{Per-Task Results}
For completeness, we report the detailed success rates for all RLBench-60 tasks in Tab.~\ref{tab:detail_rlbench} and all RoboMimic settings in Tab.~\ref{tab:detail_robomimic}. These results correspond to the aggregated metrics presented in the main paper and provide a more fine-grained comparison across tasks.

\begin{longtable}{l l c c c}
\caption{Detailed Results on RLBench-60}
\label{tab:detail_rlbench}\\

\toprule
Simplified name & Full name & DP & CoA & SegDiff \\
\midrule
\endfirsthead

\toprule
Simplified name & Full name & DP & CoA & SegDiff \\
\midrule
\endhead
lamp on & lamp\_on & 0.00 & 0.48 & 0.80 \\
remove plate & take\_plate\_off\_colored\_dish\_rack & 0.12 & 0.40 & 0.80 \\
pick up cup & pick\_up\_cup & 0.00 & 0.80 & 0.92 \\
reach target & reach\_target & 0.08 & 0.84 & 0.96 \\
push button & push\_button & 0.12 & 0.76 & 0.92 \\
open grill & open\_grill & 0.00 & 0.24 & 0.56 \\
hang frame & hang\_frame\_on\_hanger & 0.00 & 0.16 & 0.52 \\
lamp off & lamp\_off & 0.20 & 0.68 & 0.88 \\
drag to target & reach\_and\_drag & 0.28 & 0.64 & 0.88 \\
turn tap & turn\_tap & 0.32 & 0.56 & 0.84 \\
remove toilet roll & take\_toilet\_roll\_off\_stand & 0.08 & 0.56 & 0.72 \\
turn oven on & turn\_oven\_on & 0.28 & 0.36 & 0.72 \\
place plate & put\_plate\_in\_colored\_dish\_rack & 0.04 & 0.32 & 0.56 \\
move hanger & move\_hanger & 0.32 & 0.88 & 0.96 \\
open drawer & open\_drawer & 0.44 & 0.88 & 1.00 \\
phone on base & phone\_on\_base & 0.04 & 0.80 & 0.76 \\
unplug charger & unplug\_charger & 0.20 & 0.60 & 0.68 \\
take out shoes & take\_shoes\_out\_of\_box & 0.16 & 0.08 & 0.40 \\
remove frame & take\_frame\_off\_hanger & 0.24 & 0.64 & 0.72 \\
stack wine & stack\_wine & 0.56 & 0.80 & 0.96 \\
get ice & get\_ice\_from\_fridge & 0.24 & 0.60 & 0.68 \\
place knife & place\_knife\_on\_chopping\_board & 0.00 & 0.04 & 0.28 \\
open wine & open\_wine\_bottle & 0.40 & 0.36 & 0.64 \\
slide block & slide\_block\_to\_target & 0.00 & 0.52 & 0.52 \\
put bottle in fridge & put\_bottle\_in\_fridge & 0.00 & 0.28 & 0.40 \\
close grill & close\_grill & 0.68 & 0.56 & 0.88 \\
open box & open\_box & 0.32 & 0.32 & 0.56 \\
hang hanger & place\_hanger\_on\_rack & 0.00 & 0.32 & 0.40 \\
remove meat & meat\_off\_grill & 0.16 & 0.88 & 0.76 \\
open door & open\_door & 0.60 & 0.92 & 1.00 \\
remove pan lid & take\_lid\_off\_saucepan & 0.60 & 0.80 & 0.92 \\
remove umbrella & \makecell[l]{take\_umbrella\_out\_of\\\_umbrella\_stand} & 0.20 & 0.52 & 0.56 \\
put money in safe & put\_money\_in\_safe & 0.24 & 0.80 & 0.72 \\
press switch & press\_switch & 0.56 & 0.44 & 0.68 \\
water plants & water\_plants & 0.00 & 0.48 & 0.40 \\
remove from scale & take\_off\_weighing\_scales & 0.64 & 0.84 & 0.88 \\
play jenga & play\_jenga & 0.72 & 1.00 & 1.00 \\
close box & close\_box & 0.76 & 1.00 & 1.00 \\
screw nail & screw\_nail & 0.00 & 0.08 & 0.16 \\
open microwave & open\_microwave & 0.40 & 0.44 & 0.52 \\
set clock & change\_clock & 0.20 & 0.40 & 0.40 \\
take out money & take\_money\_out\_safe & 0.68 & 0.76 & 0.80 \\
hockey & hockey & 0.00 & 0.08 & 0.12 \\
close drawer & close\_drawer & 0.76 & 1.00 & 0.96 \\
basketball & basketball\_in\_hoop & 0.72 & 0.76 & 0.80 \\
seat down & toilet\_seat\_down & 0.88 & 1.00 & 1.00 \\
put groceries on & put\_groceries\_in\_cupboard & 0.00 & 0.08 & 0.08 \\
hit ball & hit\_ball\_with\_cue & 0.00 & 0.08 & 0.08 \\
sweep dust & sweep\_to\_dustpan & 1.00 & 0.92 & 1.00 \\
buzz game & beat\_the\_buzz & 0.00 & 0.36 & 0.20 \\
change channel & change\_channel & 0.00 & 0.12 & 0.08 \\
straighten rope & straighten\_rope & 0.00 & 0.00 & 0.00 \\
close fridge & close\_fridge & 0.76 & 0.92 & 0.84 \\
open washer & open\_washing\_machine & 0.60 & 0.76 & 0.64 \\
set checkers & setup\_checkers & 0.04 & 0.04 & 0.00 \\
lift block & lift\_numbered\_block & 0.08 & 0.08 & 0.04 \\
remove USB & take\_usb\_out\_of\_computer & 0.72 & 0.60 & 0.60 \\
insert USB & insert\_usb\_in\_computer & 0.88 & 0.92 & 0.84 \\
seat up & toilet\_seat\_up & 0.88 & 0.84 & 0.76 \\
\bottomrule
\end{longtable}

\begin{table}[h]
\centering
\caption{Success Rate (Mean) on 3 RoboMimic Tasks under 30/50/100/200 Demonstrations}
\begin{tabular}{@{}clccc@{}}
\toprule
Demos & Task & DP & AWE & \textbf{SegDiff} \\
\midrule
     & Lift   & \textbf{1.00} & \textbf{1.00} & \textbf{1.00} \\
30   & Can    & 0.64 & 0.62 & \textbf{0.66} \\
     & Square & 0.42 & 0.36 & \textbf{0.44} \\
\midrule
     & Lift   & \textbf{1.00} & \textbf{1.00} & \textbf{1.00} \\
50   & Can    & 0.86 & 0.77 & \textbf{0.87} \\
     & Square & 0.52 & 0.49 & \textbf{0.55} \\
\midrule
     & Lift   & \textbf{1.00} & \textbf{1.00} & \textbf{1.00} \\
100  & Can    & \textbf{0.93} & 0.87 & 0.92 \\
     & Square & 0.78 & 0.77 & \textbf{0.78} \\
\midrule
     & Lift   & \textbf{1.00} & \textbf{1.00} & \textbf{1.00} \\
200  & Can    & \textbf{0.98} & 0.96 & \textbf{0.98} \\
     & Square & \textbf{0.94} & 0.88 & 0.92 \\
\bottomrule
\end{tabular}
\label{tab:detail_robomimic}
\end{table}

\end{document}